\documentclass[sigconf]{acmart}
\usepackage{multirow} 
\AtBeginDocument{%
  }

\copyrightyear{2026}
\acmYear{2026}
\setcopyright{cc}
\setcctype{by}
\acmConference[MM '26] {Proceedings of the 34th ACM International Conference on Multimedia}{November 10--14, 2026}{Rio de Janeiro, Brazil.}
\acmBooktitle{Proceedings of the 34th ACM International Conference on Multimedia (MM '26), November 10--14, 2026, Rio de Janeiro, Brazil}
\acmISBN{979-8-4007-2213-4/2026/11}
\acmDOI{10.1145/3767308.3836342}
\begin{document}

\title{Do Protective Perturbations Really Protect Portrait Privacy under Real-world Image Transformations?}


\author{Rui-Qing Sun}
\authornote{Both authors contributed equally to this research.}
\affiliation{%
  \institution{Beijing Institute of Technology}
  \city{Beijing}
  \country{China}
}
\email{3120245673@bit.edu.cn}

\author{Xingshan Yao}
\authornotemark[1]  
\affiliation{%
  \institution{Beijing Institute of Technology}
  \city{Beijing}
  \country{China}
}
\email{3120245900@bit.edu.cn}
\author{Zhijing Wu}
\authornote{Corresponding author.}
\affiliation{%
  \institution{Beijing Institute of Technology}
  \city{Beijing}
  \country{China}
}
\email{wuzhijing.joyce@gmail.com}

\author{Tian Lan}
\affiliation{%
  \institution{Beijing Institute of Technology}
  \city{Beijing}
  \country{China}
}
\email{lantiangmftby@gmail.com}

\author{Chen-Hao Cui}
\affiliation{%
  \institution{Beijing Institute of Technology}
  \city{Beijing}
  \country{China}
}
\email{djxx208cch@163.com}

\author{Hui-Yang Zhao}
\affiliation{%
  \institution{Beijing Institute of Technology}
  \city{Beijing}
  \country{China}
}
\email{1953542921@qq.com}

\author{Jia-Ling Shi}
\affiliation{%
  \institution{Beijing Institute of Technology}
  \city{Beijing}
  \country{China}
}
\email{1120220612@bit.edu.cn}

\author{Chen Yang}
\authornotemark[2]
\affiliation{%
  \institution{Beijing Institute of Technology}
  \city{Beijing}
  \country{China}
}
\email{yangchen666@bit.edu.cn}

\author{Xian-Ling Mao}
\affiliation{%
  \institution{Beijing Institute of Technology}
  \city{Beijing}
  \country{China}
}
\email{maoxl@bit.edu.cn}

\renewcommand{\shortauthors}{Sun et al.}


\begin{abstract}
    Proactive defense methods protect portrait images from unauthorized editing or talking face generation (TFG) by introducing pixel-level protective perturbations, and have already attracted increasing attention for privacy protection.
    In real-world use, images inevitably undergo sequences of benign operations during display across devices and subsequent dissemination, such as resizing and color compression, which directly alter pixel values.
    Existing studies and robustness defenses have mainly examined individual transformations in isolation, while the effects of sequentially composed transformations remain underexplored.
    To address this gap, we systematically evaluate whether representative proactive defenses against unauthorized image editing using GANs and diffusion models, as well as unauthorized talking face generation, remain effective under sequential image transformations.
    The evaluated methods span both general purpose and portrait specific defenses and are assessed through qualitative and quantitative evaluation.
    Experimental results indicate that defense methods based on pixel-level perturbations struggle to withstand sequences of common image transformations, posing a risk of defense failure in real-world applications. 
    To further demonstrate that this vulnerability can be exploited at low cost, we introduced Transformation-Induced Purification via Region-wise Super-Resolution (TIP-RSR), a simple training-free purification framework that combines sequential image transformations with off-the-shelf restoration models.
    TIP-RSR can efficiently purify protective perturbations while preserving image fidelity and requiring substantially less computation than existing diffusion-based purification methods.
    These findings expose a practical vulnerability in current proactive portrait defenses and highlight the need to explicitly account for the effects of sequential real-world transformations when designing future protection mechanisms. Our code is publicly available at https://github.com/Richen7418/TIP-RSR.
    %
\end{abstract}

\ccsdesc[500]{Security and privacy~Privacy protections}
\ccsdesc[300]{Computing methodologies~Computer vision}

\keywords{Portrait Privacy Protection; Adversarial Purification; Super-Resolution}

\maketitle

\section{Introduction}
The rapid advancement of generative adversarial networks (GANs) and diffusion models has substantially accelerated the development of AI-generated content (AIGC) technologies, such as image editing and talking face generation (TFG)  \cite{xu2024hallo,brooks2023instructpix2pix,chen2020simswap,choi2018stargan}. While these advances demonstrate remarkable technical capabilities, they also substantially increase the risk of malicious misuse, including fraud, political manipulation, and the dissemination of misinformation.
\begin{figure}[t]
    \centering
    \includegraphics[width=\linewidth]{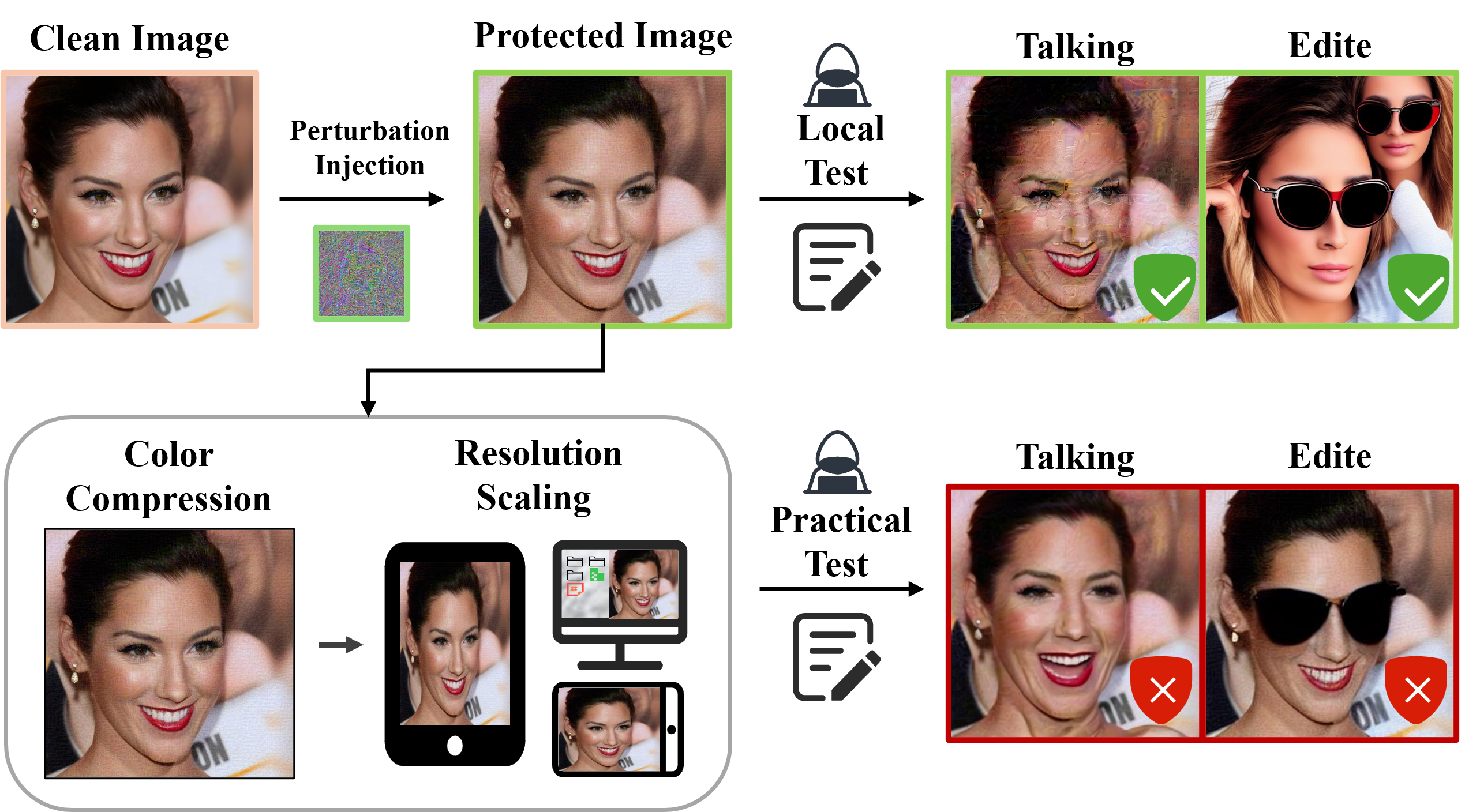}
    \caption{Impact of real-world image transformations on defense performance. Talking-face samples are generated by Hallo \cite{xu2024hallo}, while edited images are produced by InstructPix2Pix \cite{brooks2023instructpix2pix} with the prompt ``Let the person wear sunglasses.''}
    \label{fig:motivation}
\end{figure}

To mitigate these risks, a variety of defense techniques have been proposed and can be broadly categorized into passive detection and proactive defense \cite{deng2025defense_survey}. Passive detection methods attempt to identify whether facial content has been edited or forged by training generative content detectors \cite{zheng2024breaking_sem_artifacts}. In contrast, proactive defense methods proactively interfere with the generation process by injecting imperceptible protective perturbations into images, causing the synthesized outputs to degrade \cite{gan2025silence,qu2024Df-rap,wang2025facelock}. By preventing privacy abuse at the source, proactive defense has therefore attracted increasing attention. Recent studies demonstrate that such perturbations can effectively prevent generative models from modifying facial attributes or identity~\cite{jeong2025faceshield,wang2025facelock}, or from synthesizing talking portrait videos using TFG models~\cite{gan2025silence,sun2025video_TFG_defense} under controlled evaluation settings. 
Some studies also have investigated how image compression during dissemination on online social network platforms affects the effectiveness of proactive defenses \cite{qu2024Df-rap,11095758RUIP}. 

However, existing studies generally evaluate image transformations separately and therefore overlook the cumulative effects of transformation sequences. In practical use, an image may first be resized through interpolation to fit displays with different resolutions and screen dimensions, and then compressed or reencoded during transmission, sharing, or storage. Because most existing defenses are not designed to withstand such sequences and remain highly sensitive to their combined effects, even benign user actions, such as taking a screenshot for sharing or archiving, may unintentionally weaken the protection, as illustrated in Fig. \ref{fig:motivation}. Similar degradation can also result from color compression in mobile imaging pipelines, processing applied by different platforms, and conversion between storage formats, all of which irreversibly alter pixel values and image statistics. Consequently, evaluations limited to isolated transformations may substantially overestimate robustness in practical deployment and ultimately give users a false sense of security.

In this work, we systematically evaluate the effectiveness of representative proactive defenses after protected portrait images undergo common real-world transformations, either individually or in sequential combinations. Our study considers two representative application scenarios: (i) protecting portrait images against malicious facial attribute manipulation and (ii) preventing portrait images from being exploited for TFG. To ensure broad coverage, we evaluate a diverse set of defense methods with publicly available implementations for AIGC systems built on GANs and diffusion models, including both general purpose and portrait specific approaches. The evaluation covers both widely cited methods and recently proposed approaches representing the current state of the art. Extensive experiments show that protective perturbations often fail to reliably prevent portrait editing or talking face synthesis under common transformations, such as resizing and color compression, whether applied individually or in sequence. Notably, this vulnerability persists even for methods specifically \cite{qu2024Df-rap,xue2023toward_effective_protection} optimized to improve robustness to the corresponding transformations in isolation.

To further raise awareness of these risks within the community, we propose a simple yet effective training-free purification method, termed Transformation-Induced Purification via Region-wise Super-Resolution (TIP-RSR), by exploiting the above vulnerabilities. The proposed approach adopts differentiated purification strategies for facial regions and background areas in portrait images. Experimental results show that TIP-RSR can efficiently remove protective perturbations while maximally preserving the structural integrity and visual fidelity of the original image. These findings underscore the urgent need to revisit the robustness assumptions of existing protection methods and to explore more resilient solutions for mitigating privacy abuse risks posed by modern AIGC technologies.

Our contributions are summarized as follows:
\begin{itemize}
    \item We systematically evaluate the effectiveness of representative proactive defense methods, including widely used and recent approaches, under resizing and JPEG compression, both individually and in sequence.
    \item We demonstrate that common real-world transformations can significantly degrade the effectiveness of protective perturbations, causing existing methods to fail to reliably prevent portrait editing and TFG in practice.
    \item We introduce TIP-RSR, a training-free purification approach that effectively removes protective perturbations from protected images while preserving most of the original structural information, exposing previously overlooked vulnerabilities in current protection methods.
\end{itemize}

\section{Related Works}
\subsection{Talking Face Generation}
Generating realistic talking portrait videos conditioned on a given audio signal and a target portrait image is a core task in TFG, and has attracted considerable attention due to its broad practical applications. 
Early works primarily focused on synthesizing lip movements that are highly synchronized with the input audio, achieving notable success in scenarios such as video translation and dubbing \cite{wave2lip,kr2019towards}.
More recently, the development of diffusion models has substantially advanced the TFG field \cite{sora,live_avater}. The Hallo \cite{xu2024hallo} series introduces hierarchical audio-driven visual synthesis modules along with data augmentation strategies, enabling high-resolution, pose-controllable, and long-duration talking portrait video generation. Loopy \cite{jiang_loopy} exploits long-term motion dependencies to synthesize vivid talking portrait videos with subtle facial details, without relying on explicit motion constraints or predefined templates. Ditto \cite{li2025ditto} further improves diffusion-based TFG frameworks by enabling streaming processing, real-time inference, and low first-frame latency.
\subsection{Image Editing}
In the field of AIGC, image editing has been widely recognized as a task of significant practical importance. Early studies were predominantly built upon GAN-based frameworks, enabling the manipulation of specific visual attributes such as hair color, skin tone, and even facial identity \cite{choi2018stargan,chen2020simswap}.
More recently, the rapid development of multimodal large-scale models has made it possible to directly edit images according to natural language instructions. For example, InstructPix2Pix \cite{brooks2023instructpix2pix} leverages multiple large models to construct bimodal paired data, enabling instruction-following image editing guided by human-written instructions. HiDream-I1 \cite{cai2025hidream} adopts a sparse Diffusion Transformer architecture together with a dynamic mixture-of-experts design, achieving a favorable trade-off between computational efficiency and image generation quality. Qwen-Image \cite{wu2025qwen_image} further improves visual consistency and semantic coherence between edited images and their originals through comprehensive data engineering, progressive learning strategies, enhanced multi-task training paradigms, and scalable infrastructure optimizations.
While these advances in image editing and TFG substantially improve controllability, fidelity, and usability, they also exacerbate ethical and societal risks, particularly raising concerns about privacy violations and the misuse of AIGC.
\subsection{Proactive Defense}
To protect personal images from potential misuse by TFG and image editing models, several recent studies have explored perturbation-based protection by injecting imperceptible noise to disrupt the generation process. FaceLock \cite{wang2025facelock} integrates a facial recognition model as an adversary into the diffusion loop, inducing substantial visual discrepancies between edited outputs and the original images. DF-RAP \cite{qu2024Df-rap} explicitly incorporates compression operations commonly used on online
social network platforms, thereby improving the robustness of image editing defenses under practical deployment conditions. The Silencer \cite{gan2025silence} series aims to silence speech generation by applying protective perturbations to the input images, achieving effective defense against TFG methods.
Despite their effectiveness, these approaches generally overlook the fact that images may undergo sequences of benign transformations, such as resolution changes and color compression, during real-world dissemination, which can significantly undermine the reliability of such protections in practice.

\section{Evaluation of Protective Perturbation}
\label{Evaluation}
In this section, we systematically assess the robustness of representative proactive defenses under common real-world image transformations. Our evaluation covers both facial image editing and TFG, considering transformations applied individually and in sequence.
\subsection{Dataset and Selected Models}

Our evaluation focuses on two key application scenarios: facial attribute editing and TFG. To ensure that the evaluation metrics more accurately reflect the effectiveness of defense methods, we adopt the CelebA-HQ \cite{karras2018celebA-HQ} dataset, as it is explicitly used in the original papers of the evaluated methods, thereby minimizing potential interference introduced by out-of-domain data. 
Considering the substantial computational cost of large-scale evaluation, following the work of Ditto \cite{li2025ditto}, we select a subset of the CelebA-HQ dataset consisting of the first 100 facial images. For GAN-based facial attribute editing experiments, we employ DF-RAP \cite{qu2024Df-rap} and Anti-Forgery \cite{wang2022anti-forgery} as representative proactive defense methods to protect portrait images against StarGAN-based attacks \cite{choi2018stargan}. Notably, DF-RAP explicitly accounts for the influence of online social network platforms, making it particularly suitable for evaluating defense robustness under realistic dissemination conditions.
For diffusion-based experiments, we adopt two recently released, publicly available defense methods, FaceLock \cite{wang2025facelock} and FaceShield \cite{jeong2025faceshield}, to protect portrait images against attacks performed using InstructPix2Pix \cite{brooks2023instructpix2pix} and IP-Adapter \cite{ye2023ip-adapter}.
For TFG attacks, we evaluate a set of general-purpose defense methods, including AdvDM(-) \cite{xue2023toward_effective_protection}, Mist \cite{liang2023mist}, and PhotoGuard \cite{salman2023PhotoGuard}, together with the task-specific defense Silencer-I \cite{gan2025silence}. All methods are applied to protect portrait images against attacks generated by Hallo \cite{xu2024hallo}.

\subsection{Evaluation Metrics} 
To investigate how common image processing operations affect protective perturbations, we apply three representative transformations to adversarially protected samples: JPEG compression with a quality factor of 75, image resizing, and a combination of JPEG compression followed by resizing. The transformed images are then used as inputs for both image editing and TFG tasks. Following DF-RAP \cite{qu2024Df-rap}, qualitative and quantitative analyses are conducted by comparing the generated results with those obtained using undefended clean images as inputs.


Across our evaluations, we employ SSIM, PSNR, FID~\cite{heusel2017FID}, and identity similarity (ID-S) where applicable.
SSIM and PSNR measure structural and pixel-level fidelity, while FID evaluates distribution-level visual quality. ID-S measures facial identity consistency and is computed as the cosine similarity between face embeddings extracted from corresponding image pairs using a pretrained face recognition model, where a higher score indicates greater identity similarity. For image editing, we additionally report LPIPS~\cite{zhang2018Lpips} and BRISQUE~\cite{mittal2011BRIQSQUE} to assess perceptual similarity and image quality, respectively. For TFG, Sync-C~\cite{chung2016sync-c} and M-LMD~\cite{chen2019M-LMD} are further adopted to evaluate audio-visual synchronization and lip-motion consistency, respectively. 

\subsection{Experimental Results}
Experimental results show that existing defense methods struggle to remain effective under common image transformations, especially when multiple operations are applied sequentially. As shown in Tables~\ref{tab:defense_metrics} and~\ref{tab:defense_metrics_tfg}, C\&R processing makes both edited images and TFG outputs substantially closer to those generated from clean inputs. For image editing, all evaluated defenses exhibit higher SSIM and PSNR and lower LPIPS, while FID and BRISQUE vary across methods. For TFG, visual fidelity and lip motion consistency improve consistently, as reflected by higher SSIM, PSNR, and Sync-C and lower FID and M-LMD. Additional quantitative results for settings involving resize-only and JPEG-only transformations are provided in Tables \ref{tab:defense_metrics_tfg_appendix} and Table \ref{tab:defense_metrics_appendix} in the Appendix.

\begin{figure}[!htbp]
    \centering
    \includegraphics[width=\linewidth]{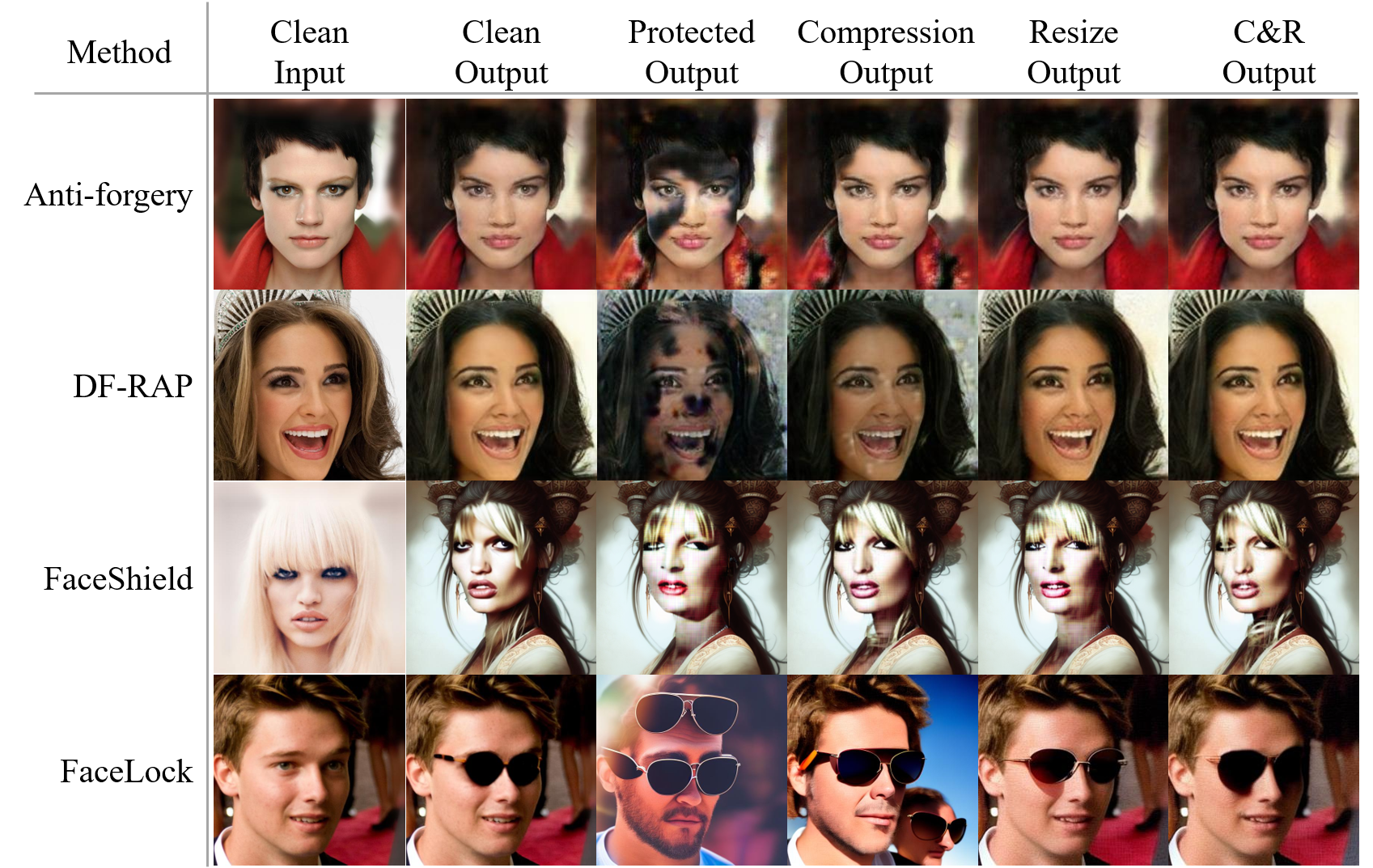}
    \caption{
    Qualitative results of image editing under different transformation settings.
    C\&R denotes JPEG compression ($Q=75$) followed by $0.5\times$ Lanczos down-sampling.
    }
    \label{fig:jr_image_edit}
\end{figure}

\begin{figure}[!htbp]
    \centering
    \includegraphics[width=\linewidth]{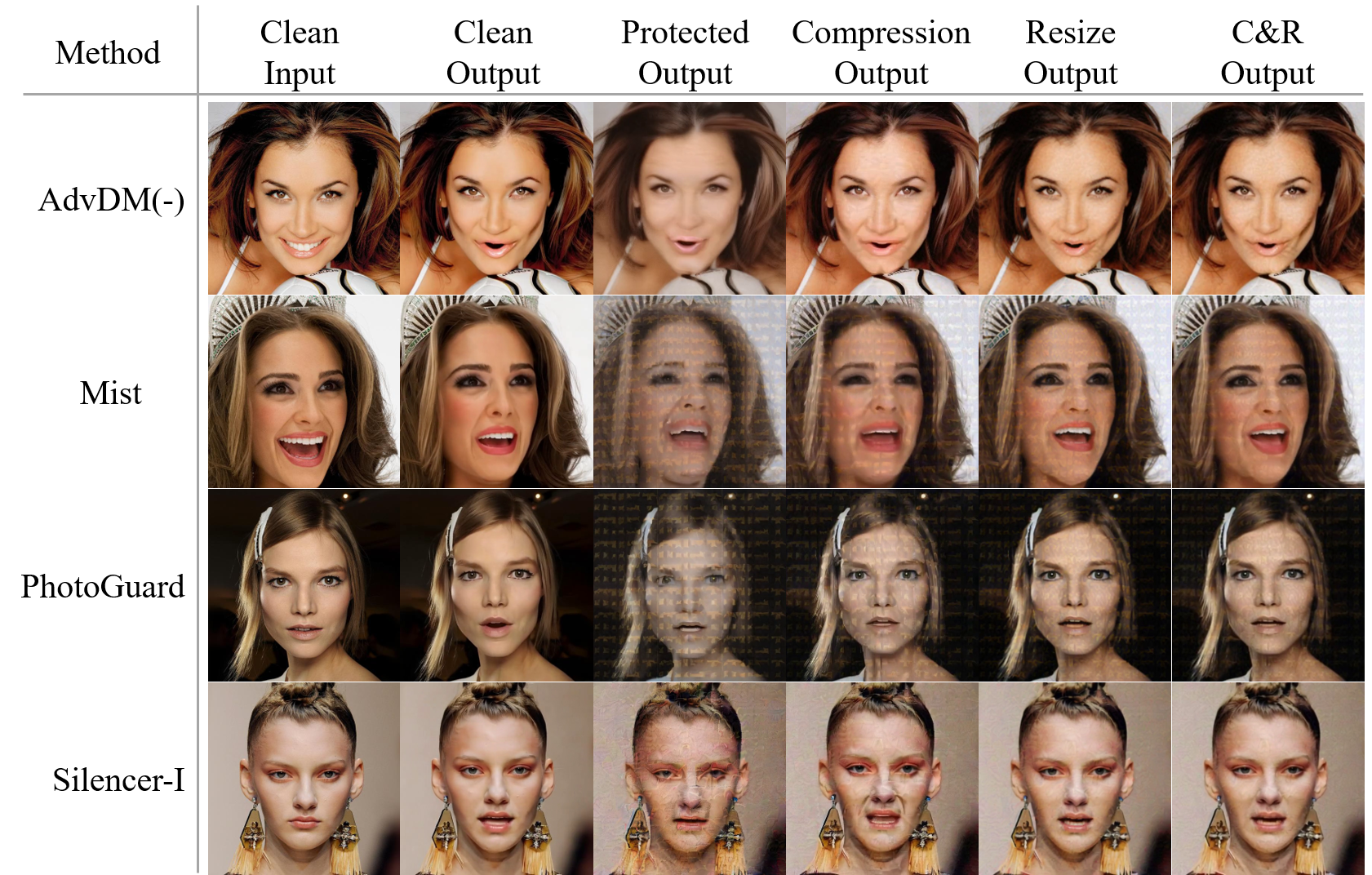}
    \caption{
    Qualitative results of TFG under different transformation settings.
    C\&R denotes JPEG compression ($Q=75$) followed by $0.5\times$ Lanczos down-sampling.
    }
    \label{fig:jr_TFG}
\end{figure}

Qualitative results show that existing defense models tend to lose their effectiveness after undergoing scale variations, and this degradation becomes more pronounced when scale changes are combined with JPEG compression (C\&R), as illustrated in Fig. \ref{fig:jr_image_edit} and Fig. \ref{fig:jr_TFG}. The evaluation results further suggest that common real-world image processing operations should be explicitly considered during model design, which otherwise may result in overestimated defense performance and a false sense of security for users. DF-RAP exhibits relatively stronger robustness to JPEG compression because its design explicitly accounts for compression operations commonly applied by online social network platforms. However, this targeted robustness does not generalize to the C\&R setting, where compression and resizing are applied sequentially.  
Such targeted modeling strategies remain inherently fragile. For instance, they struggle to defend against joint attacks under the C\&R setting, where compression and resizing are applied in combination. This observation suggests that explicitly modeling a single image transformation in isolation may not constitute a robust solution.
This behavior can be attributed to the nature of image transformations in real-world dissemination, which typically preserve perceptually salient content while altering visually less significant image details. Since many proactive defenses similarly rely on imperceptible perturbations embedded in such image details, these perturbations can be readily disrupted during transformation, thereby weakening or even nullifying the intended protection.

\begin{table}[t]
  \centering
  \footnotesize
  \setlength{\tabcolsep}{3pt}
  \caption{
  Image editing results before and after C\&R processing. C\&R denotes JPEG compression ($Q=75$) followed by $0.5\times$ Lanczos down-sampling.
  "$\uparrow$": higher is better. "$\downarrow$": lower is better.
  }
  \label{tab:defense_metrics}
  \resizebox{\linewidth}{!}{
  \begin{tabular}{c|c|ccccc}
    \toprule
    \textbf{Type} & \textbf{Method} & \textbf{SSIM}$\uparrow$ & \textbf{PSNR}$\uparrow$ & \textbf{FID}$\downarrow$ & \textbf{LPIPS}$\downarrow$ & \textbf{BRISQUE}$\downarrow$\\
    \midrule
    \multirow{4}{*}{Defense}
    & Anti-Forgery & 0.8422 & 23.94 & 39.33 & 0.1993 & 23.99 \\
    & DF-RAP       & 0.5130 & 12.93 & 139.6 & 0.4611 & 29.20 \\
    & FaceShield  & 0.8709 & 21.78 & 45.56 & 0.1026 & 27.21 \\
    & FaceLock    & 0.5202 & 16.57 & 56.34 & 0.3830 & 7.302 \\
    \midrule
    \multirow{4}{*}{C\&R}
    & Anti-Forgery & 0.8924 & 28.25 & 31.04 & 0.1600 & 28.31 \\
    & DF-RAP       & 0.8813 & 27.37 & 32.54 & 0.1640 & 28.53 \\
    & FaceShield  & 0.8835 & 22.20 & 50.95 & 0.0979 & 34.63 \\
    & FaceLock    & 0.6015 & 17.33 & 56.61 & 0.3491 & 5.213 \\
    \bottomrule
  \end{tabular}}
\end{table}

\begin{table}[t]
  \centering
  \footnotesize
  \setlength{\tabcolsep}{3pt}
  \caption{
  TFG results before and after C\&R processing.
  C\&R denotes JPEG compression ($Q=75$) followed by $0.5\times$ Lanczos down-sampling.
  }
  \label{tab:defense_metrics_tfg}
  \resizebox{\linewidth}{!}{
  \begin{tabular}{c|c|ccccc}
    \toprule
    \textbf{Type} & \textbf{Method} & \textbf{SSIM}$\uparrow$ & \textbf{PSNR}$\uparrow$ & \textbf{FID}$\downarrow$ & \textbf{Sync-C}$\uparrow$ & \textbf{M-LMD}$\downarrow$ \\
    \midrule
    \multirow{4}{*}{{Defense}}
    & AdvDM(-)    & 0.5425 & 15.98 & 110.6 & 6.773 & 18.58 \\
    & Mist        & 0.5304 & 18.39 & 277.8 & 6.049 & 19.90 \\
    & PhotoGuard  & 0.5384 & 18.25 & 272.9 & 6.037 & 17.81 \\
    & Silencer-I  & 0.4693 & 19.42 & 241.2 & 4.131 & 19.54 \\
    \midrule
    \multirow{4}{*}{C\&R}
    & AdvDM(-)    & 0.7090 & 22.63 & 75.67 & 6.889 & 9.834 \\
    & Mist        & 0.6494 & 21.77 & 125.5 & 6.841 & 10.66 \\
    & PhotoGuard  & 0.6499 & 21.85 & 127.4 & 6.801 & 10.34 \\
    & Silencer-I  & 0.6973 & 22.57 & 78.46 & 6.761 & 9.590 \\
    \bottomrule
  \end{tabular}}
\end{table}

\section{Purification Method}
In this section, we first formulate the purification objective for face manipulation and then introduce TIP-RSR, a training-free framework motivated by the sensitivity of protective perturbations to common image transformations. TIP-RSR combines transformation-induced perturbation suppression with region-wise super-resolution to restore image quality while preserving facial structure.
\subsection{Problem Formulation}
Both facial attribute editing and TFG can be formulated as face manipulation tasks. Given an original face image $x$ and a trained manipulation model $M$, the former aims to transform $x$ into a manipulated output $y$ by modifying a set of reference conditions $r$, such as lip color or target identity. In contrast, the latter generates a high-fidelity talking portrait video $y$ by conditioning on an input audio signal $a$ while preserving the identity information extracted from $x$. Accordingly, the objective of face manipulation can be described as follows:
\begin{equation}
y =
\begin{cases}
M(x, r), & \text{image editing;} \\
M(x, a), & \text{TFG.}
\end{cases}
\end{equation}
The formula can be simplified to:
\begin{equation}
    y = \mathcal{M}(x).
\end{equation}

Proactive defense methods generate adversarial examples $\hat{x} = x + \eta$ by injecting imperceptible perturbations $\eta$ into an original face image $x$. 
The perturbation $\eta$ is optimized to maximize a distance metric $D$ between the outputs of a manipulation model $\mathcal{M}$ given $x$ and $\hat{x}$, subject to an $\ell_{\infty}$-norm constraint $\|\eta\|_{\infty} \le \epsilon$ to preserve imperceptibility:
\begin{align}
\max_{\eta} \quad & D\big(\mathcal{M}(x), \mathcal{M}(x + \eta)\big), \\
\text{s.t.} \quad & \|\eta\|_{\infty} \le \epsilon .
\end{align}

Purification methods aim to remove protective perturbations from an adversarial example $\hat{x}$ and recover an image aligned with the original image $x$. The objective of purification can be formulated as minimizing the discrepancy between the purified image and the original image:
\begin{equation}
\min_{x'} \; D(x, x').
\end{equation}

Applying image transformations to an adversarial example $\hat{x}$, particularly down-sampling or compression, inevitably introduces irreversible modifications to pixel values, including the injected perturbations. As demonstrated in Section \ref{Evaluation}, such transformations can render the original adversarial perturbations ineffective. In contrast, natural images typically exhibit redundancy and strong correlations among neighboring pixels~\cite{torralba2003natural_image_categories,wang2004image_quality_assessment}, where important features and structures are often replicated across spatial regions, making critical information less susceptible to severe degradation. Although simple down-sampling alone is sufficient to invalidate most defense mechanisms, it typically cannot fully suppress residual protective perturbations and inevitably comes at the cost of reduced image resolution. Consequently, these residual artifacts may still degrade the output quality of the manipulation model $\mathcal{M}$, as further illustrated in Fig. \ref{fig:jr_image_edit} and Fig. \ref{fig:jr_TFG}.

To improve the visual quality of compressed images, a naive approach is to apply existing general super-resolution (SR) models to adversarial examples $\hat{x}$. However, most existing SR models are trained on low-resolution samples synthesized from high-resolution images via down-sampling, JPEG compression, or additive Gaussian or Poisson noise \cite{yi2025TVT,wang2021real-esrgan,chen2025ADC,lin2024Diffbir,agustsson2017Ntire}. As a result, directly applying off-the-shelf SR models to adversarial examples often fails to completely remove protective perturbations and may even introduce high-frequency artifacts on facial regions, as shown in Fig. \ref{fig:direct_SR_noisy} (Appendix). 

\subsection{TIP-RSR}

Training or fine-tuning an SR model specifically tailored to protective perturbations is a feasible alternative. Nevertheless, constructing paired datasets of clean images and adversarial examples incurs substantial computational overhead. Moreover, protective perturbations introduced by different protection methods follow distinct distributions due to their heterogeneous optimization objectives and strategies. Consequently, the training cost required for such customized SR models is typically prohibitive in practical purification scenarios.
Diffusion-based purification methods provide another line of solutions by injecting a small amount of noise during the forward diffusion process and recovering clean images through an iterative reverse denoising procedure \cite{nie2022DiffPure,pei2025freqpure}. Although these approaches avoid making explicit assumptions about the perturbation distribution, the step-wise reverse generation process introduces non-negligible computational overhead.

\begin{figure}[!htbp]
    \centering
    \includegraphics[width=\linewidth]{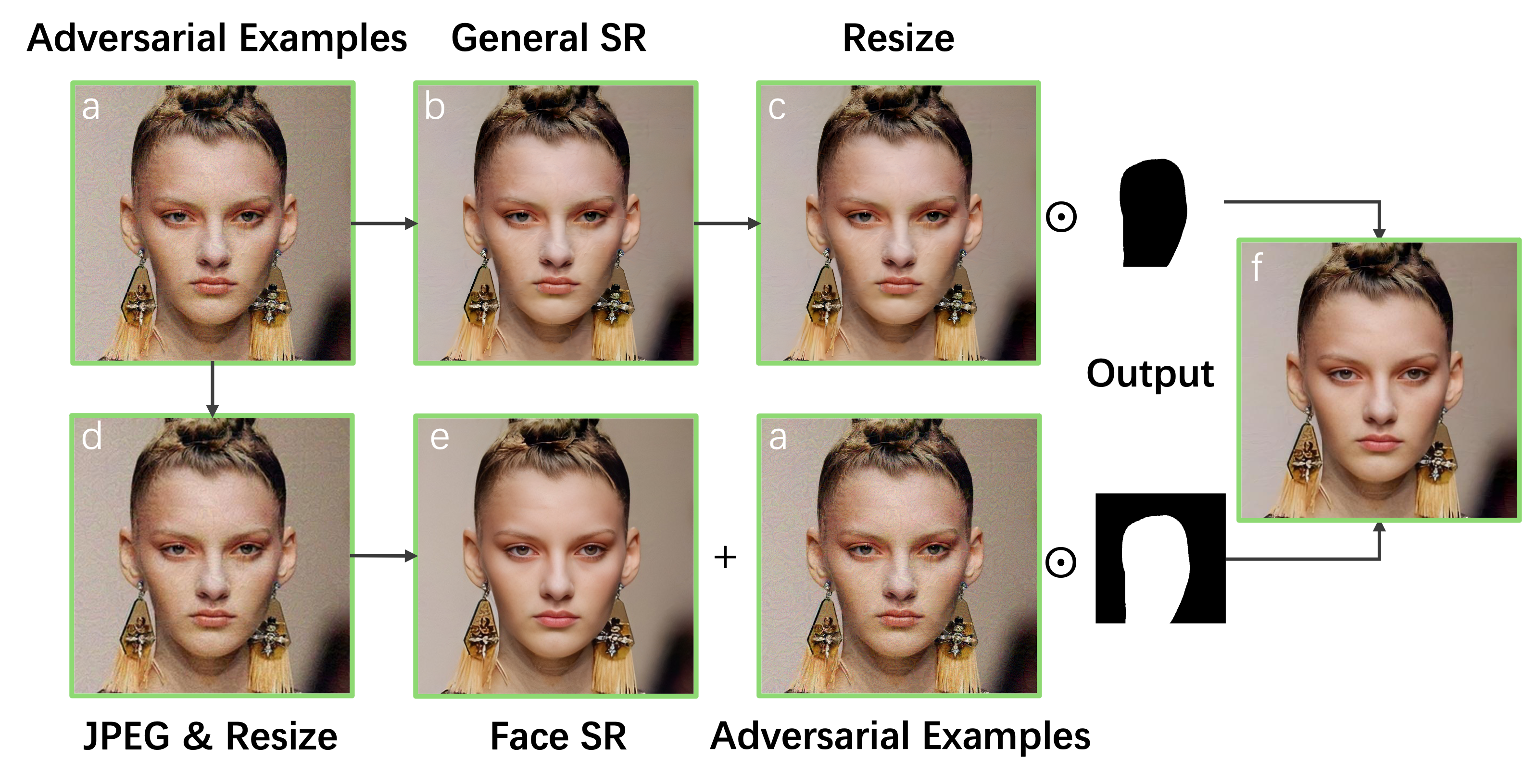}
    \caption{
    Overview of TIP-RSR. 
    }
    \label{fig:method}
\end{figure}

\begin{table*}[!htbp]
  \centering
  \footnotesize
  \setlength{\tabcolsep}{2.5pt}
  \caption{\textbf{Quality of purified protected images for image editing defenses.} 
  Bold and underlined values indicate the best and second best results, respectively.
  }
  \label{tab:image edit pure}
  \resizebox{\textwidth}{!}{
  \begin{tabular}{l|ccccc|ccccc|ccccc|ccccc}
    \toprule
    \multirow{2}{*}{Method} &
    \multicolumn{5}{c}{\textbf{Anti-Forgery}} &
    \multicolumn{5}{c}{\textbf{DF-RAP}} &
    \multicolumn{5}{c}{\textbf{FaceShield}} &
    \multicolumn{5}{c}{\textbf{FaceLock}} \\
    \cmidrule(lr){2-6} \cmidrule(lr){7-11}
    \cmidrule(lr){12-16} \cmidrule(lr){17-21}
    & SSIM$\uparrow$ & PSNR$\uparrow$ & FID$\downarrow$ & LPIPS$\downarrow$ & BRISQUE$\downarrow$
    & SSIM$\uparrow$ & PSNR$\uparrow$ & FID$\downarrow$ & LPIPS$\downarrow$ & BRISQUE$\downarrow$
    & SSIM$\uparrow$ & PSNR$\uparrow$ & FID$\downarrow$ & LPIPS$\downarrow$ & BRISQUE$\downarrow$
    & SSIM$\uparrow$ & PSNR$\uparrow$ & FID$\downarrow$ & LPIPS$\downarrow$ & BRISQUE$\downarrow$ \\
    \midrule

    DiffPure
    & 0.7879 & 28.22 & \underline{45.05} & 0.2429 & \textbf{11.88}
    & 0.7729 & 28.01 & \underline{45.57} & \underline{0.2509} & \textbf{10.09}
    & 0.7386 & 27.25 & \underline{54.85} & 0.2267 & \textbf{11.58}
    & 0.7334 & 27.21 & \underline{52.87} & 0.2329 & \textbf{11.87} \\

    GridPure
    & \underline{0.8302} & 26.26 & 63.89 & \underline{0.2255} & 32.78
    & 0.7405 & 26.06 & 95.08 & 0.3697 & 43.98
    & \underline{0.8335} & 26.93 & 55.25 & \underline{0.1455} & 44.06 
    & \underline{0.8057} & 26.75 & 58.19 & \underline{0.1589} & 28.91 \\

    FreqPure
    & 0.8195 & \underline{28.99} & 80.07 & 0.2782 & 47.47
    & \underline{0.8100} & \underline{28.86} & 80.87 & 0.2832 & 47.66
    & 0.7705 & \underline{27.95} & 87.76 & 0.2459 & 48.51
    & 0.7638 & \underline{27.84} & 90.12 & 0.2556 & 50.52 \\

    \textbf{TIP-RSR}
    & \textbf{0.9109} & \textbf{31.39} & \textbf{30.46} & \textbf{0.1282} & \underline{18.35}
    & \textbf{0.8930} & \textbf{31.15} & \textbf{35.65} & \textbf{0.1605} & \underline{20.54}
    & \textbf{0.9099} & \textbf{31.48} & \textbf{29.58} & \textbf{0.1033} & \underline{22.62}
    & \textbf{0.8532} & \textbf{29.87} & \textbf{42.89} & \textbf{0.1493} & \underline{14.85} \\
    \bottomrule
  \end{tabular}}
\end{table*}

\begin{table*}[t]
  \centering
  \footnotesize
  \setlength{\tabcolsep}{2.5pt}
  \caption{
  Quality and runtime of purified protected images for TFG defenses.
  }
  \label{tab:TFG pure}
  \resizebox{\textwidth}{!}{
  \begin{tabular}{l|ccccc|ccccc|ccccc|ccccc|c}
    \toprule
    \multirow{2}{*}{Method} &
    \multicolumn{5}{c}{\textbf{AdvDM(-)}} &
    \multicolumn{5}{c}{\textbf{Mist}} &
    \multicolumn{5}{c}{\textbf{PhotoGuard}} &
    \multicolumn{5}{c|}{\textbf{Silencer-I}} &
    \multirow{2}{*}{\textbf{Time(s)}}\\
    \cmidrule(lr){2-6} \cmidrule(lr){7-11}
    \cmidrule(lr){12-16} \cmidrule(lr){17-21}
    & SSIM$\uparrow$ & PSNR$\uparrow$ & FID$\downarrow$ & LPIPS$\downarrow$ & BRISQUE$\downarrow$
    & SSIM$\uparrow$ & PSNR$\uparrow$ & FID$\downarrow$ & LPIPS$\downarrow$ & BRISQUE$\downarrow$
    & SSIM$\uparrow$ & PSNR$\uparrow$ & FID$\downarrow$ & LPIPS$\downarrow$ & BRISQUE$\downarrow$
    & SSIM$\uparrow$ & PSNR$\uparrow$ & FID$\downarrow$ & LPIPS$\downarrow$ & BRISQUE$\downarrow$ \\
    \midrule

    DiffPure
    & 0.7280 & 27.05 & 56.95 & 0.2388 & \textbf{14.15}
    & 0.7131 & 26.34 & \textbf{66.08} & \underline{0.2673} & \textbf{13.23}
    & 0.7116 & 26.31 & \underline{67.75} & \underline{0.2704} & \textbf{13.47}
    & 0.7311 & 27.18 & \underline{55.26} & 0.2313 & \textbf{11.03} 
    & \underline{16.96} \\

    GridPure
    & \underline{0.8224} & 26.52 & \underline{56.07} & \underline{0.1605} & 54.41
    & \underline{0.7883} & 24.99 & 94.25 & 0.3008 & 34.67
    & \underline{0.7875} & 24.96 & 91.81 & 0.3036 & 36.66
    & \underline{0.8332} & 26.81 & 58.05 & \underline{0.1568} & 42.01 
    & 408.1 \\

    FreqPure
    & 0.7533 & \underline{27.54} & 91.16 & 0.2634 & 52.74
    & 0.7460 & \underline{26.84} & 109.9 & 0.3057 & 49.46
    & 0.7440 & \underline{26.78} & 112.2 & 0.3104 & 49.95
    & 0.7612 & \underline{27.86} & 88.93 & 0.2494 & 48.88  
    & 75.93 \\

    \textbf{TIP-RSR}
    & \textbf{0.8599} & \textbf{30.98} & \textbf{30.92} & \textbf{0.1185} & \underline{33.02}
    & \textbf{0.8096} & \textbf{28.77} & \underline{68.80} & \textbf{0.2581} & \underline{20.80} 
    & \textbf{0.8080} & \textbf{28.74} & \textbf{65.75} & \textbf{0.2577} & \underline{22.32}
    & \textbf{0.8671} & \textbf{30.77} & \textbf{32.54} & \textbf{0.1368} & \underline{24.24} 
    & \textbf{1.256} \\
    \bottomrule
  \end{tabular}}
\end{table*}

\begin{table*}[!ht]
  \centering
  \footnotesize
  \setlength{\tabcolsep}{2.5pt}
  \caption{
  Quality of image editing results generated from purified protected images.
  }
  \label{tab: image edit pure image generated}
  \resizebox{\textwidth}{!}{
  \begin{tabular}{l|ccccc|ccccc|ccccc|ccccc}
    \toprule
    \multirow{2}{*}{Method} &
    \multicolumn{5}{c}{\textbf{Anti-Forgery}} &
    \multicolumn{5}{c}{\textbf{DF-RAP}} &
    \multicolumn{5}{c}{\textbf{FaceShield}} &
    \multicolumn{5}{c}{\textbf{FaceLock}} \\
    \cmidrule(lr){2-6} \cmidrule(lr){7-11}
    \cmidrule(lr){12-16} \cmidrule(lr){17-21}
    & SSIM$\uparrow$ & PSNR$\uparrow$ & FID$\downarrow$ & LPIPS$\downarrow$ & ID-S$\uparrow$
    & SSIM$\uparrow$ & PSNR$\uparrow$ & FID$\downarrow$ & LPIPS$\downarrow$ & ID-S$\uparrow$
    & SSIM$\uparrow$ & PSNR$\uparrow$ & FID$\downarrow$ & LPIPS$\downarrow$ & ID-S$\uparrow$
    & SSIM$\uparrow$ & PSNR$\uparrow$ & FID$\downarrow$ & LPIPS$\downarrow$ & ID-S$\uparrow$\\
    \midrule

    DiffPure
    & 0.8120 & 26.39 & 49.27 & 0.2329 & 0.7401
    & 0.8016 & 25.86 & \underline{52.92} & 0.2430 & 0.7362
    & \underline{0.9206} & \textbf{26.27} & \underline{24.44} & \underline{0.0525} & \textbf{0.6989}
    & 0.6348 & 18.78 & 53.80 & 0.2998 & 0.5663 \\

    GridPure
    & \underline{0.8606} & 26.41 & \underline{41.11} & \underline{0.1829} & \underline{0.8452}
    & 0.5645 & 14.71 & 119.0 & 0.4478 & 0.4172
    & 0.9016 & 24.23 & 27.45 & 0.0654 & 0.5799
    & \underline{0.7093} & \underline{21.09} & \underline{40.92} & \underline{0.2246} & \underline{0.7548} \\

    FreqPure
    & 0.8412 & \underline{26.74} & 58.89 & 0.2262 & 0.7503
    & \underline{0.8351} & \underline{26.33} & 60.00 & \underline{0.2307} & \underline{0.7476}
    & 0.9005 & 24.40 & 31.60 & 0.0663 &\underline{0.6264}
    & 0.6633 & 19.35 & 70.22 & 0.3247 & 0.4827 \\

    \textbf{TIP-RSR}
    & \textbf{0.9031} & \textbf{28.96} & \textbf{25.25} & \textbf{0.1288} & \textbf{0.8657}
    & \textbf{0.8761} & \textbf{26.72} & \textbf{29.69} & \textbf{0.1579} & \textbf{0.8553}
    & \textbf{0.9323} & \underline{25.81} & \textbf{21.18} & \textbf{0.0493} & 0.6257 
    & \textbf{0.7122} & \textbf{21.70} & \textbf{34.46} & \textbf{0.2223} & \textbf{0.7761}  \\
    \bottomrule
  \end{tabular}}
\end{table*}

\begin{table*}[ht!]
  \centering
  \footnotesize
  \setlength{\tabcolsep}{2.5pt}
  \caption{
  Quality of TFG results generated from purified protected images.
  }
  \label{tab:TFG pure image generated}
  \resizebox{\textwidth}{!}{
  \begin{tabular}{l|ccccc|ccccc|ccccc|ccccc}
    \toprule
    \multirow{2}{*}{Method} &
    \multicolumn{5}{c}{\textbf{AdvDM(-)}} &
    \multicolumn{5}{c}{\textbf{Mist}} &
    \multicolumn{5}{c}{\textbf{PhotoGuard}} &
    \multicolumn{5}{c}{\textbf{Silencer-I}} \\
    \cmidrule(lr){2-6} \cmidrule(lr){7-11}
    \cmidrule(lr){12-16} \cmidrule(lr){17-21}
    & SSIM$\uparrow$ & PSNR$\uparrow$ & FID$\downarrow$ & ID-S$\uparrow$ & M-LMD$\downarrow$
    & SSIM$\uparrow$ & PSNR$\uparrow$ & FID$\downarrow$ & ID-S$\uparrow$ & M-LMD$\downarrow$
    & SSIM$\uparrow$ & PSNR$\uparrow$ & FID$\downarrow$ & ID-S$\uparrow$ & M-LMD$\downarrow$
    & SSIM$\uparrow$ & PSNR$\uparrow$ & FID$\downarrow$ & ID-S$\uparrow$ & M-LMD$\downarrow$\\
    \midrule

    DiffPure
    & 0.6759 & \underline{21.24} & 71.86 & 0.6161 & 11.13
    & 0.6607 & \underline{21.04} & \textbf{81.79} & 0.6152 & \underline{10.94}
    & \underline{0.6601} & \underline{21.03} & \textbf{83.00} & 0.6129 & \underline{10.93}
    & 0.6727 & \underline{21.21} & 70.04 & 0.6276 & 11.00 \\

    GridPure
    & \underline{0.6935} & 21.02 & \textbf{40.69} & \textbf{0.8392} & \textbf{10.30}
    & 0.6491 & 19.91 & 119.2 & \underline{0.7475} & 11.00
    & 0.6487 & 19.86 & 119.8 & \underline{0.7380} & 11.32
    & \underline{0.6742} & 21.01 & \underline{41.86} & \textbf{0.8380} & \underline{9.493} \\

    FreqPure
    & 0.6678 & 20.18 & 80.11 & 0.5973 & 14.12
    & \underline{0.6624} & 20.23 & 107.0 & 0.5993 & 13.47
    & 0.6567 & 20.07 & 110.5 & 0.5917 & 13.94
    & 0.6724 & 20.43 & 75.77 & 0.6133 & 13.57  \\

    \textbf{TIP-RSR}
    & \textbf{0.7274} & \textbf{22.57} & \underline{40.97} & \underline{0.8253} & \underline{10.31}
    & \textbf{0.6660} & \textbf{21.58} & \underline{92.00} & \textbf{0.7953} & \textbf{10.24} 
    & \textbf{0.6700} & \textbf{21.64} & \underline{89.85} & \textbf{0.7931} & \textbf{10.32} 
    & \textbf{0.7105} & \textbf{22.50} & \textbf{41.77} & \underline{0.8328} & \textbf{9.452} \\
    \bottomrule
  \end{tabular}}
\end{table*}

Motivated by the observations in Section~\ref{Evaluation}, we propose the
training-free TIP-RSR framework, as illustrated in
Fig.~\ref{fig:method}. TIP-RSR exploits the sensitivity of protective
perturbations to common image transformations while using off-the-shelf
super-resolution models to restore image quality. Specifically, we first
apply JPEG compression $J(\cdot)$ followed by down-sampling
$Down(\cdot)$ to the adversarial image $\hat{x}$, thereby suppressing
protective perturbations, as shown in Fig.~\ref{fig:method}(d). A
face-specific SR model is then applied to restore facial details.
Following GridPure~\cite{zhao2024can}, the restored result is combined
with a small proportion of the adversarial image to preserve
fine-grained structural information. Meanwhile, a general-purpose SR
model is applied to $\hat{x}$ and subsequently resized to the target
resolution to provide a structure-preserving restoration of the
remaining image content, as shown in Fig.~\ref{fig:method}(b,c).
Finally, the outputs of the two restoration branches are combined using
a semantic face mask to obtain the purified image, as illustrated in
Fig.~\ref{fig:method}(f). The overall purification process is formulated
as follows:

\begin{equation}
x_f = (1- \lambda) \cdot \mathrm{SR}_f\!\left(\mathrm{Down}(J(\hat{x}))\right) + \lambda \cdot \hat{x},
\end{equation}
\begin{equation}
x_g = \mathrm{Down}\!\left(\mathrm{SR}_g(\hat{x})\right),
\end{equation}
\begin{equation}
x' = \mathrm{Mask} \odot x_f + (1 - \mathrm{Mask}) \odot x_g .
\end{equation}
where $\mathrm{Mask}$ denotes a semantic segmentation mask used to distinguish facial regions from background regions, $\mathrm{SR}_f(\cdot)$ and $\mathrm{SR}_g(\cdot)$ represent the face-specific and general-purpose SR models, respectively, $x_f$ and $x_g$ denote their corresponding outputs, and $x'$ denotes the final purified image.


Following common practice, we adopt Lanczos interpolation for image resizing and set the JPEG quality factor to 75. The weighting parameter $\lambda$ is fixed to 0.2. The segmentation mask $\mathrm{Mask}$ is generated using BiSeNet~\cite{yu2018BiSeNet}. To balance purification effectiveness and computational efficiency, we employ a lightweight general SR model, ESRGAN~\cite{wang2018esrgan}, for background regions and a lightweight face-specific SR model, GFPGAN~\cite{wang2021GFP-GAN}, for facial regions. 
All model parameters are adopted directly from publicly available projects, and no additional training or fine-tuning is performed.
A comparison with SOTA super-resolution models is provided in Appendix.

\subsection{Experimental Results}
\subsubsection{Settings and Baselines}
The experiments were conducted on a server running Ubuntu 22.04.4 LTS, equipped with two Intel(R) Xeon(R) Gold 6133 CPUs clocked at 2.50 GHz and one NVIDIA GeForce RTX 4090 GPU with 24 GB of memory. 
We select existing publicly available purification methods, including DiffPure \cite{nie2022DiffPure}, GridPure \cite{zhao2024can}, and FreqPure \cite{pei2025freqpure}, as diffusion-based baselines. Comparisons with recent super-resolution models are provided in the appendix. We perform quantitative evaluation using the metrics introduced in Section~\ref{Evaluation}.

\subsubsection{Performance Analysis of TIP-RSR}
\begin{figure*}[!htbp]
    \centering
    \small
    \includegraphics[width=\linewidth]{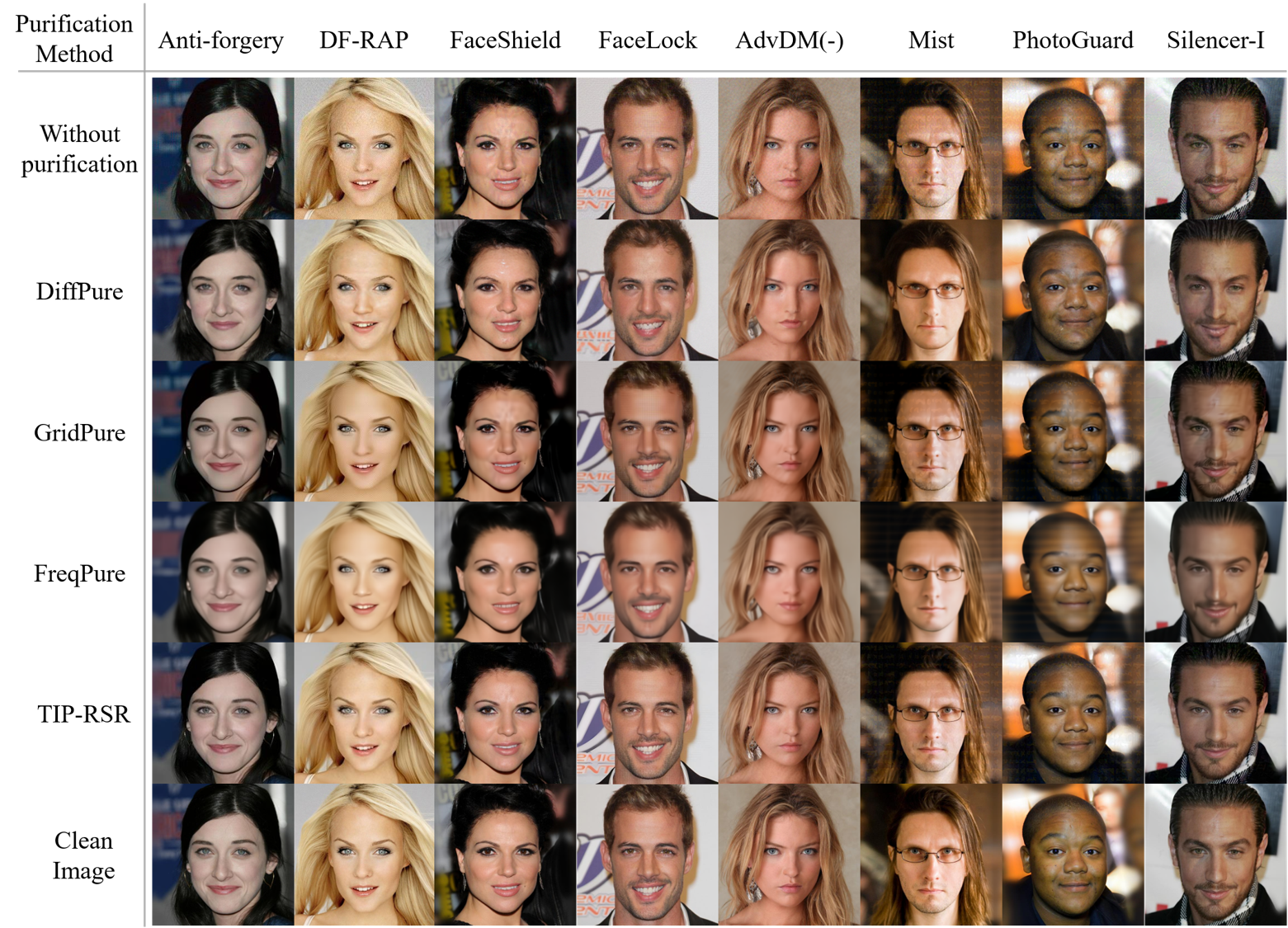}
    \caption{Qualitative comparison of purified images.}
    \label{fig:pure_img}
\end{figure*}

Experimental results demonstrate that TIP-RSR consistently outperforms existing purification methods across a wide range of evaluation metrics. As shown in Tables \ref{tab:image edit pure} and \ref{tab:TFG pure}, TIP-RSR achieves the best performance among nearly all defense settings in terms of SSIM, PSNR, FID, and LPIPS. These results indicate that TIP-RSR is able to effectively suppress adversarial perturbations while largely preserving structural fidelity and perceptual quality of the input images. It is worth noting that TIP-RSR performs worse than DiffPure on the BRISQUE metric. This behavior can be attributed to the fusion of super-resolved images with adversarial samples at a fixed ratio, which introduces a mild distribution shift from natural images. Nevertheless, qualitative results confirm that this design leads to improved perceptual quality, as illustrated in Fig.~\ref{fig:pure_img}. In contrast, images purified by DiffPure tend to appear overly smooth, resulting in noticeable loss of fine-grained details.

\begin{figure*}[t]
    \centering
    \includegraphics[width=\linewidth]{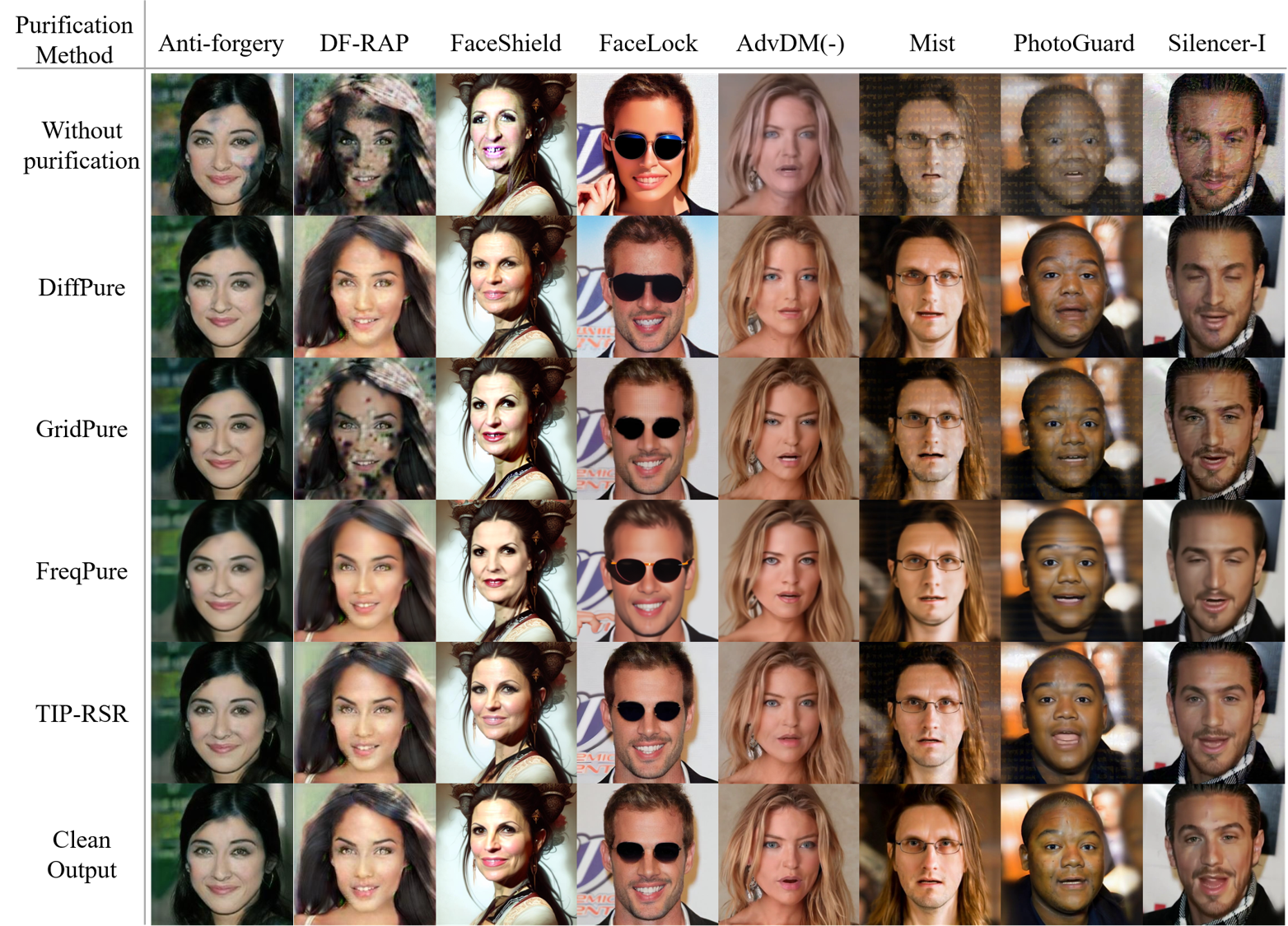}
    \caption{Qualitative results of image editing and talking face generation after image purification.}
    \label{fig:pure_generate}
\end{figure*}

The primary goal of image purification is to suppress defense-oriented perturbations while preserving the visual and identity information required by downstream generative models. As shown in Tables~\ref{tab: image edit pure image generated} and~\ref{tab:TFG pure image generated}, TIP-RSR achieves a favorable trade-off between perturbation removal, visual fidelity, and identity preservation across both image editing and TFG tasks.
For image editing, TIP-RSR achieves the best SSIM scores across all four defense settings and the best PSNR scores in three out of four cases. It also consistently obtains the lowest FID and LPIPS values, indicating that the purified images provide more faithful visual guidance for subsequent editing. Moreover, TIP-RSR achieves the highest ID-S on Anti-Forgery, DF-RAP, and FaceLock, while remaining competitive on FaceShield, demonstrating that the purification process largely preserves facial identity information in addition to improving visual quality.
For TFG, TIP-RSR consistently achieves the best SSIM and PSNR scores across all evaluated defenses, indicating improved visual fidelity of the generated video frames. It also achieves the best or second-best performance in FID and ID-S across all settings, showing that the purified portraits retain both distribution-level visual quality and identity consistency. In terms of motion quality, TIP-RSR obtains the best M-LMD results on Mist, PhotoGuard, and Silencer-I, and the second-best result on AdvDM(-), further demonstrating that effective purification does not compromise lip-motion consistency. Overall, these results suggest that TIP-RSR can reliably weaken protective perturbations while preserving the visual and identity cues required for downstream generation.
Moreover, the results highlight that purification models lacking explicit facial structural priors may struggle to fully eliminate protective perturbations. As illustrated in Fig.~\ref{fig:pure_generate}, although GridPure yields visually appealing results for some defense settings, it fails to effectively remove perturbations introduced by DF-RAP, resulting in persistently degraded editing quality.

Finally, we evaluate the computational efficiency of TIP-RSR. 
Compared with the evaluated purification baselines, TIP-RSR exhibits substantially lower processing cost. As shown in Table~\ref{tab:TFG pure}, TIP-RSR requires only 1.256 seconds on average to process a single image, making it approximately 13.5× faster than DiffPure. This efficiency stems from its feed-forward restoration pipeline, which avoids the iterative or multi-step denoising procedures used by diffusion-based baselines. 
Such efficiency highlights the strong potential of TIP-RSR for real-time applications. More importantly, the low purification cost further underscores the necessity of accounting for common real-world image transformations when designing defense models, as neglecting these factors may significantly overestimate their practical robustness.

\subsubsection{Ablation Study}
We conduct an ablation study to validate the effectiveness of our proposed TIP-RSR pipeline. Specifically, we compare it against three core baseline components:
(1) ESRGAN applied directly to the adversarial image;
(2) GFPGAN applied to the C\&R version of the adversarial image; 
(3) ESRGAN+GFPGAN, a mask-based fusion of the two outputs.

\begin{table}[htbp]
\centering
\caption{
Ablation study of TIP-RSR. The full pipeline achieves a favorable trade-off across metrics on both FaceLock and Silencer-I.
}
\label{tab:generalization}
\resizebox{\linewidth}{!}{
\begin{tabular}{c|cccc|cccc}
\toprule
\multirow{2}{*}{Method} & 
\multicolumn{4}{c}{FaceLock} & 
\multicolumn{4}{c}{Silencer-I} \\
\cmidrule(lr){2-5} \cmidrule(lr){6-9}&
SSIM$\uparrow$ & 
PSNR$\uparrow$ & 
FID$\downarrow$ & 
LPIPS$\downarrow$ & 
SSIM$\uparrow$ & 
PSNR$\uparrow$ & 
FID$\downarrow$ & 
LPIPS$\downarrow$\\
\midrule
ESRGAN
& 0.7860 & 28.55 & 65.25 & 0.1768
& \underline{0.8635} & \textbf{31.09} & 51.23 & 0.1673\\

GFPGAN
& 0.8193 & 29.31 & \textbf{37.32} & \textbf{0.1395}
& 0.8253 & 29.18 & \textbf{32.52} & \textbf{0.1273} \\

ESRGAN+GFPGAN
& \underline{0.8467} & \underline{29.49} & 46.33 & 0.1549
& 0.8610 & 30.29 & 35.59 & 0.1413 \\

TIP-RSR
& \textbf{0.8532} & \textbf{29.87} & \underline{42.89} & \underline{0.1493}
& \textbf{0.8671} & \underline{30.77} & \underline{32.54} & \underline{0.1368}\\
\bottomrule
\end{tabular}}
\end{table}

As shown in Table \ref{tab:generalization}, our full TIP-RSR method consistently achieves the best or second-best performance across all key metrics—including SSIM, PSNR, FID, and LPIPS—on both the FaceLock and Silencer-I adversarial defenses. This demonstrates that TIP-RSR effectively integrates and surpasses the individual contributions of its constituent modules.
Critically, TIP-RSR outperforms the ESRGAN+GFPGAN baseline across all reported metrics on both defenses, although both approaches combine outputs from the two SR branches. The improvement mainly comes from the fusion strategy in the face branch, which retains a controlled proportion of information from the protected image to preserve fine grained details before the final semantic mask fusion. This helps preserve high-frequency structural cues that may be lost after JPEG compression and resizing, yielding a better balance between perturbation suppression and detail fidelity. These results validate the proposed fusion design over a straightforward combination of the two SR outputs using a semantic mask.

\section{Conclusion}
In this work, we evaluate proactive defenses for portrait images under individual transformations and their sequential combinations, using image editing and TFG as representative attacks. We show that JPEG compression and resizing, especially in sequence, can substantially weaken protective perturbations and restore generation quality, exposing a practical robustness gap. We further introduce TIP-RSR, a purification framework that requires no training and combines transformation-induced perturbation suppression with region-wise super resolution. TIP-RSR removes diverse protective perturbations while preserving visual structure and identity information, and is approximately 13.5$\times$ faster than DiffPure. Our findings call for defenses that remain reliable throughout realistic image processing pipelines.

\begin{acks}
This work was supported by the National Key Research and Development Program of China (No. 2024YFF0908200), the National Natural Science Foundation of China (No. 62572056, 62302040), the Beijing Institute of Technology Research Fund Program for Young Scholars, and the Young Elite Scientists Sponsorship Program of the Beijing High Innovation Plan (No. 20250798).
\end{acks}




\bibliographystyle{ACM-Reference-Format}
\bibliography{references}

@inproceedings{wave2lip,
  title={A lip sync expert is all you need for speech to lip generation in the wild},
  author={Prajwal, KR and Mukhopadhyay, Rudrabha and Namboodiri, Vinay P and Jawahar, CV},
  booktitle={Proceedings of the 28th ACM international conference on multimedia},
  pages={484--492},
  year={2020}
}

@inproceedings{kr2019towards,
  title={Towards automatic face-to-face translation},
  author={KR, Prajwal and Mukhopadhyay, Rudrabha and Philip, Jerin and Jha, Abhishek and Namboodiri, Vinay and Jawahar, CV},
  booktitle={Proceedings of the 27th ACM international conference on multimedia},
  pages={1428--1436},
  year={2019}
}

@article{sora,
  title={Sora: A review on background, technology, limitations, and opportunities of large vision models},
  author={Liu, Yixin and Zhang, Kai and Li, Yuan and Yan, Zhiling and Gao, Chujie and Chen, Ruoxi and Yuan, Zhengqing and Huang, Yue and Sun, Hanchi and Gao, Jianfeng and others},
  journal={arXiv preprint arXiv:2402.17177},
  year={2024}
}

@article{live_avater,
  title={Live avatar: Streaming real-time audio-driven avatar generation with infinite length},
  author={Huang, Yubo and Guo, Hailong and Wu, Fangtai and Zhang, Shifeng and Huang, Shijie and Gan, Qijun and Liu, Lin and Zhao, Sirui and Chen, Enhong and Liu, Jiaming and others},
  journal={arXiv preprint arXiv:2512.04677},
  year={2025}
}

@article{xu2024hallo,
  title={Hallo: Hierarchical audio-driven visual synthesis for portrait image animation},
  author={Xu, Mingwang and Li, Hui and Su, Qingkun and Shang, Hanlin and Zhang, Liwei and Liu, Ce and Wang, Jingdong and Yao, Yao and Zhu, Siyu},
  journal={arXiv preprint arXiv:2406.08801},
  year={2024}
}

@inproceedings{jiang_loopy,
  author    = "Jiang, Jingyao and Liang, Chen and Yang, Jing and Lin, Guosheng and Zhong, Tao and Zheng, Yao",
  title     = "Loopy: Taming Audio-Driven Portrait Avatar with Long-Term Motion Dependency",
  booktitle = "Proceedings of the Thirteenth International Conference on Learning Representations",
  year      = "2025"
}

@inproceedings{li2025ditto,
  title={Ditto: Motion-space diffusion for controllable realtime talking head synthesis},
  author={Li, Tianqi and Zheng, Ruobing and Yang, Minghui and Chen, Jingdong and Yang, Ming},
  booktitle={Proceedings of the 33rd ACM International Conference on Multimedia},
  pages={9704--9713},
  year={2025}
}

@inproceedings{chen2020simswap,
  title={Simswap: An efficient framework for high fidelity face swapping},
  author={Chen, Renwang and Chen, Xuanhong and Ni, Bingbing and Ge, Yanhao},
  booktitle={Proceedings of the 28th ACM international conference on multimedia},
  pages={2003--2011},
  year={2020}
}

@inproceedings{choi2018stargan,
  title={Stargan: Unified generative adversarial networks for multi-domain image-to-image translation},
  author={Choi, Yunjey and Choi, Minje and Kim, Munyoung and Ha, Jung-Woo and Kim, Sunghun and Choo, Jaegul},
  booktitle={Proceedings of the IEEE conference on computer vision and pattern recognition},
  pages={8789--8797},
  year={2018}
}

@inproceedings{brooks2023instructpix2pix,
  title={Instructpix2pix: Learning to follow image editing instructions},
  author={Brooks, Tim and Holynski, Aleksander and Efros, Alexei A},
  booktitle={Proceedings of the IEEE/CVF conference on computer vision and pattern recognition},
  pages={18392--18402},
  year={2023}
}

@inproceedings{cai2025hidream,
  title={HiDream-I1: An Open-Source High-Efficient Image Generative Foundation Model},
  author={Cai, Qi and Li, Yehao and Pan, Yingwei and Yao, Ting and Mei, Tao},
  booktitle={Proceedings of the 33rd ACM International Conference on Multimedia},
  pages={13636--13639},
  year={2025}
}

@article{wu2025qwen_image,
  title={Qwen-image technical report},
  author={Wu, Chenfei and Li, Jiahao and Zhou, Jingren and Lin, Junyang and Gao, Kaiyuan and Yan, Kun and Yin, Sheng-ming and Bai, Shuai and Xu, Xiao and Chen, Yilei and others},
  journal={arXiv preprint arXiv:2508.02324},
  year={2025}
}

@inproceedings{wang2025facelock,
  title={Edit away and my face will not stay: Personal biometric defense against malicious generative editing},
  author={Wang, Hanhui and Zhang, Yihua and Bai, Ruizheng and Zhao, Yue and Liu, Sijia and Tu, Zhengzhong},
  booktitle={Proceedings of the Computer Vision and Pattern Recognition Conference},
  pages={23806--23816},
  year={2025}
}

@article{qu2024Df-rap,
  title={Df-rap: A robust adversarial perturbation for defending against deepfakes in real-world social network scenarios},
  author={Qu, Zuomin and Xi, Zuping and Lu, Wei and Luo, Xiangyang and Wang, Qian and Li, Bin},
  journal={IEEE Transactions on Information Forensics and Security},
  volume={19},
  pages={3943--3957},
  year={2024},
  publisher={IEEE}
}

@inproceedings{gan2025silence,
  title={Silence is Golden: Leveraging Adversarial Examples to Nullify Audio Control in LDM-based Talking-Head Generation},
  author={Gan, Yuan and Miao, Jiaxu and Wang, Yunze and Yang, Yi},
  booktitle={Proceedings of the Computer Vision and Pattern Recognition Conference},
  pages={13434--13444},
  year={2025}
}

@article{zheng2024breaking_sem_artifacts,
  title={Breaking semantic artifacts for generalized ai-generated image detection},
  author={Zheng, Chende and Lin, Chenhao and Zhao, Zhengyu and Wang, Hang and Guo, Xu and Liu, Shuai and Shen, Chao},
  journal={Advances in Neural Information Processing Systems},
  volume={37},
  pages={59570--59596},
  year={2024}
}

@article{torralba2003natural_image_categories,
  title={Statistics of natural image categories},
  author={Torralba, Antonio and Oliva, Aude},
  journal={Network: computation in neural systems},
  volume={14},
  number={3},
  pages={391},
  year={2003},
  publisher={IOP Publishing}
}

@article{wang2004image_quality_assessment,
  title={Image quality assessment: from error visibility to structural similarity},
  author={Wang, Zhou and Bovik, Alan C and Sheikh, Hamid R and Simoncelli, Eero P},
  journal={IEEE transactions on image processing},
  volume={13},
  number={4},
  pages={600--612},
  year={2004},
  publisher={IEEE}
}

@inproceedings{wang2021GFP-GAN,
  title={Towards real-world blind face restoration with generative facial prior},
  author={Wang, Xintao and Li, Yu and Zhang, Honglun and Shan, Ying},
  booktitle={Proceedings of the IEEE/CVF conference on computer vision and pattern recognition},
  pages={9168--9178},
  year={2021}
}

@inproceedings{wang2018esrgan,
  title={Esrgan: Enhanced super-resolution generative adversarial networks},
  author={Wang, Xintao and Yu, Ke and Wu, Shixiang and Gu, Jinjin and Liu, Yihao and Dong, Chao and Qiao, Yu and Change Loy, Chen},
  booktitle={Proceedings of the European conference on computer vision (ECCV) workshops},
  pages={0--0},
  year={2018}
}

@inproceedings{yi2025TVT,
  title={Fine-structure preserved real-world image super-resolution via transfer vae training},
  author={Yi, Qiaosi and Li, Shuai and Wu, Rongyuan and Sun, Lingchen and Wu, Yuhui and Zhang, Lei},
  booktitle={Proceedings of the IEEE/CVF international conference on computer vision},
  pages={12415--12426},
  year={2025}
}

@inproceedings{yu2018BiSeNet,
  title={Bisenet: Bilateral segmentation network for real-time semantic segmentation},
  author={Yu, Changqian and Wang, Jingbo and Peng, Chao and Gao, Changxin and Yu, Gang and Sang, Nong},
  booktitle={Proceedings of the European conference on computer vision (ECCV)},
  pages={325--341},
  year={2018}
}

@inproceedings{nie2022DiffPure,
  title={Diffusion Models for Adversarial Purification},
  author={Nie, Weili and Guo, Brandon and Huang, Yujia and Xiao, Chaowei and Vahdat, Arash and Anandkumar, Animashree},
  booktitle={International Conference on Machine Learning},
  pages={16805--16827},
  year={2022},
  organization={PMLR}
}

@inproceedings{pei2025freqpure,
title={Diffusion-based Adversarial Purification from the Perspective of the Frequency Domain},
author={Gaozheng Pei and Ke Ma and Yingfei Sun and Qianqian Xu and Qingming Huang},
booktitle={Forty-second International Conference on Machine Learning},
year={2025},
url={https://openreview.net/forum?id=Bm706VlAtU}
}

@inproceedings{wang2021real-esrgan,
  title={Real-esrgan: Training real-world blind super-resolution with pure synthetic data},
  author={Wang, Xintao and Xie, Liangbin and Dong, Chao and Shan, Ying},
  booktitle={Proceedings of the IEEE/CVF international conference on computer vision},
  pages={1905--1914},
  year={2021}
}

@inproceedings{chen2025ADC,
  title={Adversarial diffusion compression for real-world image super-resolution},
  author={Chen, Bin and Li, Gehui and Wu, Rongyuan and Zhang, Xindong and Chen, Jie and Zhang, Jian and Zhang, Lei},
  booktitle={Proceedings of the Computer Vision and Pattern Recognition Conference},
  pages={28208--28220},
  year={2025}
}

@inproceedings{lin2024Diffbir,
  title={Diffbir: Toward blind image restoration with generative diffusion prior},
  author={Lin, Xinqi and He, Jingwen and Chen, Ziyan and Lyu, Zhaoyang and Dai, Bo and Yu, Fanghua and Qiao, Yu and Ouyang, Wanli and Dong, Chao},
  booktitle={European conference on computer vision},
  pages={430--448},
  year={2024},
  organization={Springer}
}

@inproceedings{agustsson2017Ntire,
  title={Ntire 2017 challenge on single image super-resolution: Dataset and study},
  author={Agustsson, Eirikur and Timofte, Radu},
  booktitle={Proceedings of the IEEE conference on computer vision and pattern recognition workshops},
  pages={126--135},
  year={2017}
}

@ARTICLE{11095758RUIP,
  author={Zhang, Yibo and Lin, Weiguo and Tian, Zhihong and Min, Geyong and Xu, Junfeng and Xu, Yikun},
  journal={IEEE Transactions on Information Forensics and Security}, 
  title={Robust and Unstigmatized Imperceptible Perturbations for Rendering Face Manipulation Ineffective}, 
  year={2025},
  volume={20},
  number={},
  pages={7966-7981},
  doi={10.1109/TIFS.2025.3592565}}

@article{deng2025defense_survey,
  title={A survey of defenses against ai-generated visual media: Detection, disruption, and authentication},
  author={Deng, Jingyi and Lin, Chenhao and Zhao, Zhengyu and Liu, Shuai and Peng, Zhe and Wang, Qian and Shen, Chao},
  journal={ACM Computing Surveys},
  volume={58},
  number={5},
  pages={1--35},
  year={2025},
  publisher={ACM New York, NY}
}

@article{sun2025video_TFG_defense,
  title={Efficient and Robust Video Defense Framework against 3D-field Personalized Talking Face},
  author={Sun, Rui-qing and Yao, Xingshan and Lan, Tian and Zhao, Hui-Yang and Shi, Jia-Ling and Cui, Chen-Hao and Wu, Zhijing and Yang, Chen and Mao, Xian-Ling},
  journal={arXiv preprint arXiv:2512.21019},
  year={2025}
}

@inproceedings{jeong2025faceshield,
  title={Faceshield: Defending facial image against deepfake threats},
  author={Jeong, Jaehwan and In, Sumin and Kim, Sieun and Shin, Hannie and Jeong, Jongheon and Yoon, Sang Ho and Chung, Jaewook and Kim, Sangpil},
  booktitle={Proceedings of the IEEE/CVF International Conference on Computer Vision},
  pages={10364--10374},
  year={2025}
}

@inproceedings{karras2018celebA-HQ,
  title={Progressive Growing of GANs for Improved Quality, Stability, and Variation},
  author={Karras, Tero and Aila, Timo and Laine, Samuli and Lehtinen, Jaakko},
  booktitle={International Conference on Learning Representations},
  year={2018}
}

@inproceedings{xue2023toward_effective_protection,
  title={Toward effective protection against diffusion-based mimicry through score distillation},
  author={Xue, Haotian and Liang, Chumeng and Wu, Xiaoyu and Chen, Yongxin},
  booktitle={The Twelfth International Conference on Learning Representations},
  year={2023}
}

@inproceedings{salman2023PhotoGuard,
  title={Raising the Cost of Malicious AI-Powered Image Editing},
  author={Salman, Hadi and Khaddaj, Alaa and Leclerc, Guillaume and Ilyas, Andrew and Madry, Aleksander},
  booktitle={International Conference on Machine Learning},
  pages={29894--29918},
  year={2023},
  organization={PMLR}
}

@article{liang2023mist,
  title={Mist: Towards improved adversarial examples for diffusion models},
  author={Liang, Chumeng and Wu, Xiaoyu},
  journal={arXiv preprint arXiv:2305.12683},
  year={2023}
}

@article{wang2022anti-forgery,
  title={Anti-forgery: Towards a stealthy and robust deepfake disruption attack via adversarial perceptual-aware perturbations},
  author={Wang, Run and Huang, Ziheng and Chen, Zhikai and Liu, Li and Chen, Jing and Wang, Lina},
  journal={arXiv preprint arXiv:2206.00477},
  year={2022}
}

@article{heusel2017FID,
  title={Gans trained by a two time-scale update rule converge to a local nash equilibrium},
  author={Heusel, Martin and Ramsauer, Hubert and Unterthiner, Thomas and Nessler, Bernhard and Hochreiter, Sepp},
  journal={Advances in neural information processing systems},
  volume={30},
  year={2017}
}

@inproceedings{zhang2018Lpips,
  title={The unreasonable effectiveness of deep features as a perceptual metric},
  author={Zhang, Richard and Isola, Phillip and Efros, Alexei A and Shechtman, Eli and Wang, Oliver},
  booktitle={Proceedings of the IEEE conference on computer vision and pattern recognition},
  pages={586--595},
  year={2018}
}

@inproceedings{mittal2011BRIQSQUE,
  title={Blind/referenceless image spatial quality evaluator},
  author={Mittal, Anish and Moorthy, Anush K and Bovik, Alan C},
  booktitle={2011 conference record of the forty fifth asilomar conference on signals, systems and computers (ASILOMAR)},
  pages={723--727},
  year={2011},
  organization={IEEE}
}

@inproceedings{chung2016sync-c,
  title={Out of time: automated lip sync in the wild},
  author={Chung, Joon Son and Zisserman, Andrew},
  booktitle={Asian conference on computer vision},
  pages={251--263},
  year={2016},
  organization={Springer}
}

@inproceedings{chen2019M-LMD,
  title={Hierarchical cross-modal talking face generation with dynamic pixel-wise loss},
  author={Chen, Lele and Maddox, Ross K and Duan, Zhiyao and Xu, Chenliang},
  booktitle={Proceedings of the IEEE/CVF conference on computer vision and pattern recognition},
  pages={7832--7841},
  year={2019}
}

@inproceedings{zhao2024can,
  title={Can protective perturbation safeguard personal data from being exploited by stable diffusion?},
  author={Zhao, Zhengyue and Duan, Jinhao and Xu, Kaidi and Wang, Chenan and Zhang, Rui and Du, Zidong and Guo, Qi and Hu, Xing},
  booktitle={Proceedings of the IEEE/CVF Conference on Computer Vision and Pattern Recognition},
  pages={24398--24407},
  year={2024}
}

@article{ye2023ip-adapter,
  title={IP-Adapter: Text Compatible Image Prompt Adapter for Text-to-Image Diffusion Models},
  author={Ye, Hu and Zhang, Jun and Liu, Sibo and Han, Xiao and Yang, Wei},
  booktitle={arXiv preprint arxiv:2308.06721},
  year={2023}
}

@inproceedings{wfen,
  title={Efficient face super-resolution via wavelet-based feature enhancement network},
  author={Li, Wenjie and Guo, Heng and Liu, Xuannan and Liang, Kongming and Hu, Jiani and Ma, Zhanyu and Guo, Jun},
  booktitle={Proceedings of the 32nd ACM international conference on multimedia},
  pages={4515--4523},
  year={2024}
}
\balance

\appendix
\clearpage
\label{Appendix}


\section{Protective Perturbations Degrade under Real-world Image Transformations}

To further analyze the robustness of proactive defenses under practical deployment conditions, we separately evaluate two common post-defense image transformations: JPEG compression and image resizing. In this setting, \textit{Defense} denotes directly using the protected image as input for generation, while \textit{JPEG(Q=75)} and \textit{Resizing(0.5$\times$)} denote applying JPEG compression or resizing to the protected image before feeding it into the downstream generator.

\begin{table}[!hb]
  \centering
  \footnotesize
  \setlength{\tabcolsep}{3pt}
  \caption{Effect of JPEG compression and resizing applied after protection on generated image quality in image editing tasks.}
  \label{tab:defense_metrics_appendix}
  \resizebox{\linewidth}{!}{
  \begin{tabular}{c|c|ccccc}
    \toprule
    \textbf{Setting} & \textbf{Method} & \textbf{SSIM} $\uparrow$& \textbf{PSNR}$\uparrow$ & \textbf{FID}$\downarrow$ & \textbf{LPIPS}$\downarrow$ & \textbf{BRISQUE}$\downarrow$\\
    \midrule
    \multirow{4}{*}{Defense}
    & Anti-Forgery & 0.8422 & 23.94 & 39.33 & 0.1993 & 23.99 \\
    & DF-RAP       & 0.5130 & 12.93 & 139.6 & 0.4611 & 29.20 \\
    & FaceShield  & 0.8709 & 21.78 & 45.56 & 0.1026 & 27.21 \\
    & FaceLock    & 0.5202 & 16.57 & 56.34 & 0.3830 & 7.302 \\
    \midrule
    \multirow{4}{*}{JPEG(Q=75)}
    & Anti-Forgery & 0.8939 & 28.04 & 26.83 & 0.1504 & 24.31 \\
    & DF-RAP       & 0.6762 & 16.35 & 72.21 & 0.3056 & 27.77 \\
    & FaceShield  & 0.8943 & 22.76 & 37.63 & 0.0868 & 35.59 \\
    & FaceLock    & 0.5192 & 16.27 & 57.08 & 0.3905 & 7.527 \\
    \midrule
    \multirow{4}{*}{Resizing(0.5$\times$)}
    & Anti-Forgery & 0.8928 & 28.30 & 31.66 & 0.1598 & 28.31 \\
    & DF-RAP       & 0.8526 & 25.62 & 42.55 & 0.2023 & 27.56 \\
    & FaceShield  & 0.8717 & 21.83 & 45.67 & 0.1021 & 27.60 \\
    & FaceLock    & 0.6246 & 20.21 & 50.59 & 0.3234 & 4.167 \\
    \bottomrule
  \end{tabular}}
\end{table}

\begin{table}[!hbp]
  \centering
  \footnotesize
  \setlength{\tabcolsep}{3pt}
  \caption{Effect of JPEG compression and resizing applied after protection on generated video quality in talking face generation (TFG) tasks.}
  \label{tab:defense_metrics_tfg_appendix}
  \resizebox{\linewidth}{!}{
  \begin{tabular}{c|c|ccccc}
    \toprule
    \textbf{Setting} & \textbf{Method} & \textbf{SSIM}$\uparrow$ & \textbf{PSNR}$\uparrow$ & \textbf{FID}$\downarrow$ & \textbf{Sync}$\uparrow$ & \textbf{M-LMD}$\downarrow$ \\
    \midrule
    \multirow{4}{*}{Defense}
    & AdvDM(-)    & 0.5425 & 15.98 & 110.6 & 6.773 & 18.58 \\
    & Mist        & 0.5304 & 18.39 & 277.8 & 6.049 & 19.90 \\
    & PhotoGuard  & 0.5384 & 18.25 & 272.9 & 6.037 & 17.81 \\
    & Silencer-I  & 0.4693 & 19.42 & 241.2 & 4.131 & 19.54 \\
    \midrule
    \multirow{4}{*}{JPEG(Q=75)}
    & AdvDM(-)    & 0.6448 & 20.09 & 60.25 & 6.747 & 13.72 \\
    & Mist        & 0.5734 & 20.47 & 178.3 & 6.470 & 12.82 \\
    & PhotoGuard  & 0.5788 & 20.48 & 177.2 & 6.468 & 13.01 \\
    & Silencer-I  & 0.6048 & 21.11 & 109.1 & 6.342 & 10.87 \\
    \midrule
    \multirow{4}{*}{Resizing(0.5$\times$)}
    & AdvDM(-)    & 0.7002 & 22.43 & 89.75 & 6.899 & 10.46 \\
    & Mist        & 0.6283 & 21.30 & 155.2 & 6.737 & 11.04 \\
    & PhotoGuard  & 0.6296 & 21.31 & 155.4 & 6.750 & 11.07 \\
    & Silencer-I  & 0.6703 & 21.69 & 94.20 & 6.684 & 10.65 \\
    \bottomrule
  \end{tabular}}
\end{table}

As shown in Tables~\ref{tab:defense_metrics_appendix} and  \ref{tab:defense_metrics_tfg_appendix}, applying JPEG compression or resizing  to protected images generally improves the generation quality in both image  editing and talking face generation (TFG) tasks. 
Specifically, SSIM and PSNR  tend to increase, while FID, LPIPS, and M-LMD decrease for most defense  methods, indicating that the generated results become closer to those obtained  from clean inputs. However, the extent of degradation varies across different  defenses. Some methods maintain relatively stronger robustness against a  single transformation, while others experience a substantial reduction in  defensive effectiveness.

These results demonstrate that common post-processing operations can weaken perturbation-based proactive defenses even when applied individually. JPEG  compression and resizing modify pixel-level perturbations while largely  preserving the semantic content of the image, making them capable of partially  suppressing protective perturbations. More importantly, these observations   motivate the investigation of sequential transformation pipelines, where multiple common operations may jointly amplify perturbation degradation and further compromise the robustness of existing defenses.



\section{Quantitative Results on Multiple Defenses}

To further evaluate the cross-defense effectiveness of TIP-RSR, we test it against four additional proactive defense methods with different optimization strategies and perturbation patterns. The goal of this experiment is not merely to improve image quality, but to investigate whether TIP-RSR can generalize across different protective perturbations and consistently recover higher-quality images from defended inputs.

As shown in Table~\ref{tab:multi_defense}, after applying TIP-RSR, the purified images exhibit improved reconstruction quality across all evaluated defenses, as reflected by higher SSIM and PSNR and lower FID and LPIPS in most cases. These improvements indicate that TIP-RSR can effectively suppress protective perturbations introduced by different defense mechanisms while recovering visual information closer to that of clean images.

These results demonstrate that TIP-RSR is not tailored to a specific defense mechanism, but instead shows strong transferability across multiple perturbation-based protection methods. Although different defenses employ distinct optimization objectives and perturbation patterns, TIP-RSR consistently reduces their impact on image quality after purification. This cross-defense generalization highlights the potential vulnerability of existing proactive defenses when facing simple yet effective transformation-based purification pipelines.

\begin{table}[htbp]
\centering
\caption{Cross-defense evaluation of TIP-RSR on multiple perturbation-based proactive defense methods.}
\label{tab:multi_defense}
\resizebox{1.0\linewidth}{!}{
\begin{tabular}{c|c|ccccc}
\toprule
\textbf{Setting} & \textbf{Method} & \textbf{SSIM}$\uparrow$ & \textbf{PSNR}$\uparrow$ & \textbf{FID}$\downarrow$ & \textbf{LPIPS}$\downarrow$ & \textbf{BRISQUE}$\downarrow$ \\
\midrule
\multirow{4}{*}{Defense} 
& AdvDM(+)\cite{xue2023toward_effective_protection} 
& 0.5878 & 27.36 & 125.1 & 0.4029 & 17.42 \\
& SDS(-)\cite{xue2023toward_effective_protection}
& 0.6172 & 28.56 & 56.12 & 0.2108 & 51.99 \\
& SDS(+)\cite{xue2023toward_effective_protection}
& 0.6180 & 27.61 & 132.1 & 0.4055 & 13.73 \\
& Silencer-II\cite{gan2025silence}
& 0.6098 & 25.91 & 183.1 & 0.4360 & 12.04 \\
\midrule
\multirow{4}{*}{TIP-RSR} 
& AdvDM(+)\cite{xue2023toward_effective_protection}
& 0.7844 & 27.97 & 90.28 & 0.2570 & 8.808 \\
& SDS(-)\cite{xue2023toward_effective_protection}
& 0.8551 & 31.04 & 40.59 & 0.1496 & 34.13 \\
& SDS(+)\cite{xue2023toward_effective_protection}
& 0.7607 & 27.22 & 106.3 & 0.2712 & 7.837 \\
& Silencer-II\cite{gan2025silence}
& 0.7165 & 26.62 & 122.5 & 0.3348 & 25.31 \\
\bottomrule
\end{tabular}}
\end{table}

\section{Motivation for Region-wise Fusion in TIP-RSR}

The design of TIP-RSR is motivated by the different restoration requirements of facial and non-facial regions. GFPGAN provides a useful example of such a region-wise restoration strategy: its practical inference pipeline focuses on restoring facial regions with a face-specific restoration model, while a general-purpose background upsampler can be used to enhance non-facial regions. The restored faces are then pasted back onto the upsampled background. This design is particularly suitable for portrait images, where facial regions contain semantically important details that benefit from face-specific priors, whereas background regions are better preserved by general image restoration methods.

\begin{figure}[!htbp]
    \small
    \centering
    \includegraphics[width=0.9\linewidth]{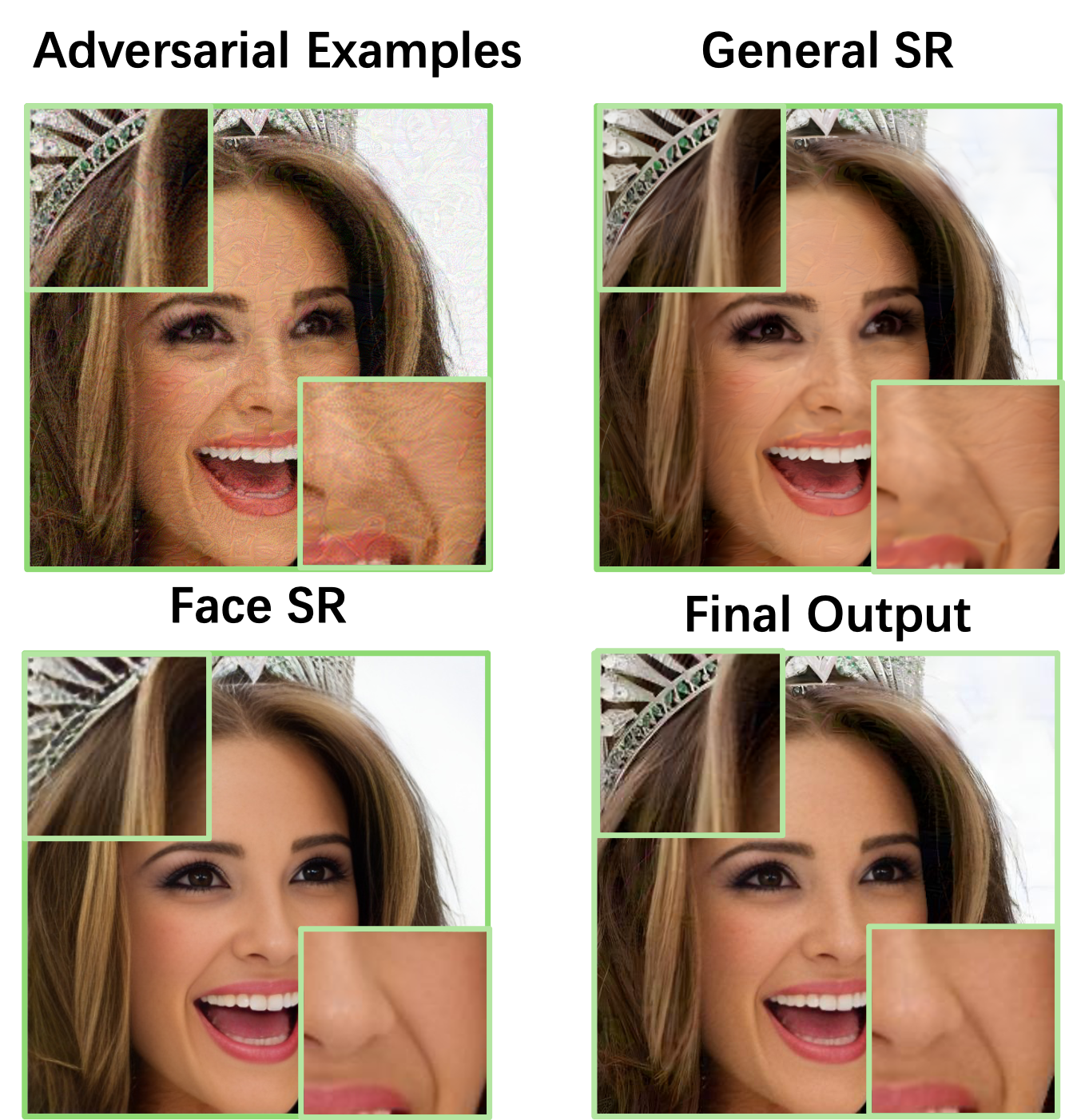}
    \caption{
    \textbf{Motivation for the region-wise fusion strategy in TIP-RSR.}
    The GFPGAN-based pipeline restores facial details after transformation-induced perturbation suppression but may degrade non-facial structures. Direct ESRGAN restoration better preserves background details, motivating the proposed face/background region-wise fusion.
    }
    \label{fig:method_motivation}
\end{figure}

However, directly applying the GFPGAN pipeline to our protected images reveals an important limitation. In TIP-RSR, JPEG compression followed by down-sampling is first applied to the protected image to suppress protective perturbations. Although these transformations are effective at weakening the perturbations, they also inevitably remove high-frequency structural information. When the transformed image is subsequently processed by the GFPGAN pipeline, the background restoration branch operates on an already degraded image, resulting in the loss of fine-grained structural details in non-facial regions. As illustrated in Fig.~\ref{fig:method_motivation}, structures and textures in the background can become noticeably blurred after this process, even though the facial region is effectively restored.

This observation motivates us to treat facial and non-facial regions differently. For facial regions, we first apply JPEG compression and down-sampling before GFPGAN, exploiting the sensitivity of protective perturbations to these transformations and the strong facial priors of GFPGAN to restore facial details. In contrast, background perturbations are often manifested as unnatural high-frequency artifacts. Applying a general-purpose SR model directly to the original protected image can effectively suppress such artifacts while better preserving the structural information of the background. Therefore, we employ ESRGAN as the general SR branch and apply it directly to the protected image without the preliminary JPEG compression and down-sampling.
Finally, we combine the two restoration results according to a semantic face mask, using the face-specific branch for facial regions and the general SR branch for non-facial regions. In this way, TIP-RSR avoids unnecessarily degrading background structures while retaining the strong facial restoration capability of GFPGAN. The resulting region-wise fusion provides a better balance between protective perturbation removal and structural fidelity, which motivates the overall design of TIP-RSR.

\section{Analysis of Super-Resolution Model Selection}

To justify the model design in TIP-RSR, we compare several representative super-resolution (SR) models for adversarial image purification under the Silencer-I defense setting. The purpose of this experiment is not to identify the best-performing SR model on generic restoration benchmarks, but to investigate which models are more suitable for our purification objective. Different from conventional SR tasks, adversarial image purification requires a balance between recovering visual details, mitigating residual perturbation artifacts, preserving identity-related facial structures, and maintaining computational efficiency for downstream generation.

As shown in Table~\ref{tab:SR}, SR models with stronger restoration capability on standard benchmarks do not necessarily achieve better performance in the adversarial purification setting. Although TVT, a recently proposed SR model, achieves competitive reconstruction quality, its inference time exceeds 27 seconds per image, which introduces substantial computational overhead for our training-free purification pipeline. In contrast, ESRGAN and GFPGAN provide a better balance between restoration quality and efficiency. Specifically, ESRGAN achieves the highest SSIM and PSNR scores, demonstrating stronger structural preservation capability, while GFPGAN obtains the best FID and LPIPS results, indicating superior perceptual quality and facial detail restoration.

These results provide the empirical basis for our SR model selection. Rather than selecting models solely based on recency or performance on generic SR benchmarks, we prioritize their behavior under adversarial purification and computational efficiency. ESRGAN and GFPGAN exhibit complementary advantages in the evaluated metrics while maintaining substantially lower inference costs than more computationally intensive alternatives such as TVT. We therefore adopt ESRGAN and GFPGAN in TIP-RSR, providing an effective balance between restoration quality and computational efficiency.

\begin{table}[h]
\centering
\caption{Comparison of representative super-resolution models for adversarial image purification under the Silencer-I defense setting. Bold and underline indicate the best and second-best results, respectively.}
\label{tab:SR}
\resizebox{1.0\linewidth}{!}{
\begin{tabular}{c|cccccc}
\toprule
\textbf{SR Model} & \textbf{SSIM}$\uparrow$ & \textbf{PSNR}$\uparrow$ & \textbf{FID}$\downarrow$ & \textbf{LPIPS}$\downarrow$ & \textbf{BRISQUE}$\downarrow$ & \textbf{Time(s)}$\downarrow$ \\
\midrule
AdcSR\cite{chen2025ADC} 
& 0.7559 & 28.53 & 55.21 & 0.2033 & \textbf{14.87} & 4.95 \\

WFEN\cite{wfen}      
& 0.7547 & 26.68 & 111.4 & 0.3120 & 45.98 & \textbf{0.40} \\

TVT\cite{yi2025TVT}         
& 0.8198 & \underline{30.09} & \underline{49.68} & 0.1844 & \underline{19.58} & 27.7 \\

ESRGAN\cite{wang2018esrgan}                                     
& \textbf{0.8635} & \textbf{31.09} & 51.23 & \underline{0.1673} & 21.15 & 0.78 \\

GFPGAN\cite{wang2021GFP-GAN}                    
& \underline{0.8288} & 29.63 & \textbf{38.27} & \textbf{0.1477} & 27.49 & \underline{0.71} \\

\bottomrule
\end{tabular}}
\end{table}

\section{Comparison with Vision Foundation Models (VFMs) for Image Purification}

To further investigate whether large-scale vision foundation models (VFMs) provide additional advantages for adversarial image purification, we compare TIP-RSR with several representative VFMs accessed through their official APIs. Unlike TIP-RSR, these models are developed for general-purpose image generation or editing and benefit from large-scale training data and model capacity. This comparison aims to examine whether such generative capabilities translate into improved purification performance under protective perturbations.

As shown in Table~\ref{tab:vfm_results}, TIP-RSR achieves competitive or superior purification performance compared with VFMs while requiring substantially fewer parameters and no additional training. Specifically, TIP-RSR obtains the highest SSIM and PSNR scores, demonstrating stronger structural fidelity after purification. Meanwhile, VFMs such as Seedream4.5 achieve competitive perceptual quality in terms of FID and LPIPS, suggesting that large-scale generative models can produce visually plausible outputs. However, foundation models are powerful generators, but purification is not equivalent to generation. While VFMs benefit from strong generative priors and can synthesize visually plausible content, adversarial purification requires preserving original structures and identity-related information rather than generating visually realistic alternatives.Therefore, large-scale generative capability does not necessarily translate into superior purification performance. 

Moreover, compared with API-based VFMs that rely on large-scale model architectures and external inference resources, TIP-RSR only requires lightweight off-the-shelf restoration models without additional training or optimization. With approximately 0.1B parameters, TIP-RSR provides a more efficient and accessible purification solution while maintaining competitive reconstruction quality.

\begin{figure*}[!ht]
    \centering
    \includegraphics[width=0.9\linewidth]{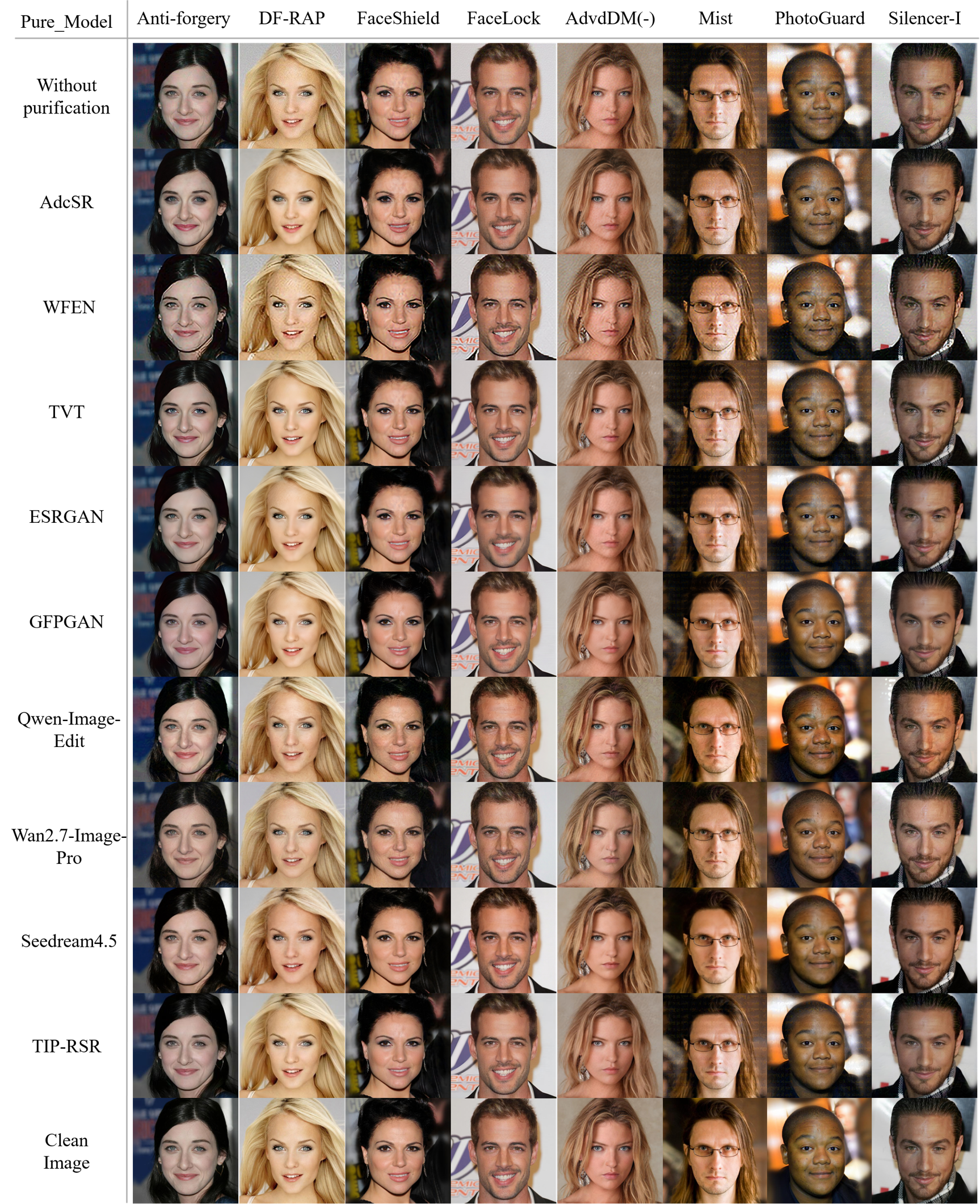}
    \caption{Qualitative comparison of TIP-RSR and vision foundation models (VFMs) for adversarial image purification. TIP-RSR better preserves structural details and facial information after purification.}
    \label{fig:direct_SR_noisy}
\end{figure*}

Qualitative comparisons in Fig.~\ref{fig:direct_SR_noisy} further demonstrate this advantage. Compared with VFM-based purification, TIP-RSR better preserves fine-grained structures and facial details, producing outputs closer to the original content. These observations suggest that effective purification does not necessarily require large-scale generative models, and lightweight, carefully designed restoration pipelines can already pose practical challenges to existing perturbation-based protection methods.

Overall, these findings highlight a fundamental vulnerability of current proactive defense methods: their robustness can be significantly weakened by simple, efficient, and widely accessible purification techniques, raising concerns about their reliability under real-world deployment conditions.

\begin{table}[htbp]
\centering
\caption{Comparison of TIP-RSR with vision foundation models (VFMs) for adversarial image purification. }
\label{tab:vfm_results}
\resizebox{1.0\linewidth}{!}{
\begin{tabular}{c|cccccc}
\toprule
\textbf{VFMs} & \textbf{SSIM}$\uparrow$ & \textbf{PSNR}$\uparrow$ & \textbf{FID}$\downarrow$ & \textbf{LPIPS}$\downarrow$ & \textbf{BRISQUE}$\downarrow$ & \textbf{Params(B)} \\
\midrule
Qwen-Image-Edit & 0.6719 & 23.26 & 54.41 & 0.2334 & 12.53 & 20 \\
Wan2.7-Image-Pro & 0.7824 & 26.13 & \underline{38.91} & \underline{0.1838} & \textbf{5.45} & -- \\
Seedream4.5 & \underline{0.8391} & \underline{27.58} & \textbf{25.81} & \textbf{0.1338} & \underline{11.06} & $\sim$12 \\
TIP-RSR & \textbf{0.8671} & \textbf{30.77} & 32.54 & 0.1368 & 24.24 & $\sim$0.1 \\
\bottomrule
\end{tabular}}
\end{table}

\end{document}